\documentclass[sigconf,natbib=true,nonacm]{acmart}

\renewcommand\footnotetextcopyrightpermission[1]{}
\usepackage{array}
\usepackage{listings}
\newcolumntype{P}[1]{>{\centering\arraybackslash}p{#1}}
\usepackage{pifont}
\usepackage{algorithm}
\usepackage{algpseudocode}
\usepackage{multirow}
\usepackage{textcomp}
\usepackage{tcolorbox}
\tcbuselibrary{breakable}

\AtBeginDocument{%
  }

\acmYear{2026}
\copyrightyear{2026}

\setcopyright{none}

\begin{document}

\title{A Hybrid Knowledge-Grounded Framework for Safety and Traceability in Prescription Verification}


\author{Yichi Zhu}
\email{Y30241060@mail.ecust.edu.cn}
\affiliation{%
  \institution{School of Information Science and Engineering, East China University of Science and Technology}
  \city{Shanghai}
  \country{China}
}

\author{Kan Ling}
\email{Y30241065@mail.ecust.edu.cn}
\affiliation{%
  \institution{School of Information Science and Engineering, East China University of Science and Technology}
  \city{Shanghai}
  \country{China}
}

\author{Xu Liu}
\email{Y30241061@mail.ecust.edu.cn}
\affiliation{%
  \institution{School of Information Science and Engineering, East China University of Science and Technology}
  \city{Shanghai}
  \country{China}
}

\author{Hengrun Zhang}
\authornote{Corresponding author.}
\email{zhanghengrun@ecust.edu.cn}
\affiliation{%
  \institution{School of Information Science and Engineering, East China University of Science and Technology}
  \city{Shanghai}
  \country{China}
}

\author{Huiqun Yu}
\authornotemark[1]
\email{yhq@ecust.edu.cn}
\affiliation{%
  \institution{School of Information Science and Engineering, East China University of Science and Technology}
  \city{Shanghai}
  \country{China}
}

\author{Guisheng Fan}
\authornotemark[1]
\email{gsfan@ecust.edu.cn}
\affiliation{%
  \institution{School of Information Science and Engineering, East China University of Science and Technology}
  \city{Shanghai}
  \country{China}
}

\renewcommand{\shortauthors}{Zhu et al.}

\begin{abstract}
    Medication errors pose a significant threat to patient safety, making pharmacist verification (PV) a critical, yet heavily burdened, final safeguard. The direct application of Large Language Models (LLMs) to this zero-tolerance domain is untenable due to their inherent factual unreliability, lack of traceability, and weakness in complex reasoning. To address these challenges, we introduce \textit{PharmGraph-Auditor}, a novel system designed for safe and evidence-grounded prescription auditing. The core of our system is a trustworthy \textit{Hybrid Pharmaceutical Knowledge Base (HPKB)}, implemented under the \textit{Virtual Knowledge Graph (VKG)} paradigm. This architecture strategically unifies a relational component for set constraint satisfaction and a graph component for topological reasoning via a rigorous mapping layer. To construct this HPKB, we propose the \textit{Iterative Schema Refinement (ISR)} algorithm, a framework that enables the co-evolution of both graph and relational schemas from medical texts. For auditing, we introduce the \textit{KB-grounded Chain of Verification (CoV)}, a new reasoning paradigm that transforms the LLM from an unreliable generator into a transparent reasoning engine. CoV decomposes the audit task into a sequence of verifiable queries against the HPKB, generating hybrid query plans to retrieve evidence from the most appropriate data store. Experimental results demonstrate robust knowledge extraction capabilities and show promises of using PharmGraph-Auditor to enable pharmacists to achieve safer and faster prescription verification.
\end{abstract}



\keywords{Prescription Auditing, Large Language Models, Hybrid Knowledge Base, Chain of Verification, Information Retrieval, Explainable AI}

\maketitle

\section{Introduction}
\label{sec:intro}

The advent of Large Language Models (LLMs) presents a promising solution to medication-related tasks. With their ability to process vast amounts of unstructured text, LLMs are well-suited for digesting information from prescribing information, medical literature, and clinical guidelines. Currently, most works focus on simple diagnostic tasks or general biomedical fact-checking \cite{sridharan2024unlocking, lu2025doctorrag, barone2025combining}, which are primarily based on medical Question-Answer pairs. However, the credibility of prescriptions in clinical settings still heavily relies on manual checks, a process that is increasingly strained by the complexity of modern pharmaceutical evidence.

Medication errors, such as incorrect dosages and adverse interactions, represent a persistent healthcare challenge, contributing to tens of thousands of adverse events and even death annually \cite{bates1995incidence,aspden2007preventing,pais2024large}. The final safeguard against these risks is \textbf{Pharmacist Verification (PV)}, where pharmacists meticulously scrutinize prescriptions to intercept potential mistakes. However, this manual defense is under increasing strain as pharmacists must navigate a data deluge, including an ever-expanding pharmacopeia, complex patient histories, and evolving clinical guidelines. This cognitive overload, often compounded by high workloads, increases the risk that critical details are overlooked \cite{gorbach2015frequency}, leading to severe patient harm. Consequently, there is a pressing need for intelligent systems capable of augmenting the pharmacist's expertise with a systematic and evidence-based safety layer.

\begin{figure}[t]
	\centering
	\includegraphics[width=0.45\textwidth]{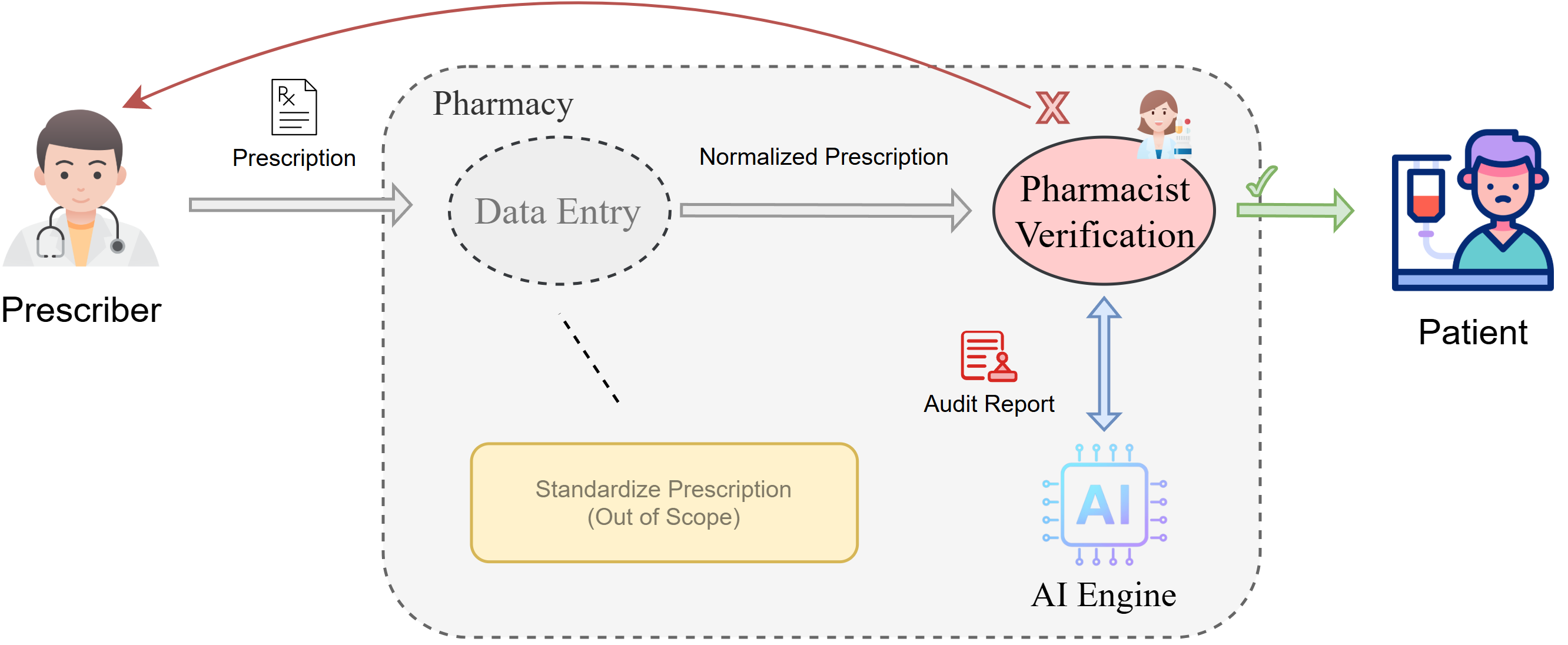}
	\caption{The workflow and the role of PharmGraph-Auditor.}
	\label{fig:workflow}
\end{figure}

Although some works aim to prevent medication errors, they rarely target the critical \textbf{PV} stage directly. As illustrated in Fig. \ref{fig:workflow}, systems like Pais et al. \cite{pais2024large} employ AI modules primarily for prescription standardization during Data Entry. In contrast, our work addresses the final verification step, where direct LLM application is untenable due to fundamental limitations: (1) \textbf{Factual Unreliability:} LLMs are prone to ``hallucination'', generating plausible but incorrect information—unacceptable where patient safety is at stake. (2) \textbf{Lack of Traceability:} Knowledge is opaquely encoded within model parameters, making it impossible to trace conclusions back to source documents. This violates evidence-based medicine, as untraceable recommendations are inherently untrustworthy. (3) \textbf{Weakness in Complex Reasoning:} Auditing requires multi-hop reasoning to connect disparate facts (e.g., patient renal function, drug properties, and dosage guidelines). LLMs struggle to perform such structured reasoning reliably without a factual scaffold.

To bridge these gaps, we introduce \textbf{PharmGraph-Auditor}. Our system constructs and queries a \textbf{Hybrid Pharmaceutical Knowledge Base (HPKB)} based on the \textbf{Virtual Knowledge Graph (VKG)} paradigm. Guided by our \textbf{Knowledge Stratification Framework}, we recognize that pharmaceutical data inherently requires dual modeling: a Relational Component ($\mathcal{R}$) to handle strict ``Constraints'' (e.g., dosage limits, numerical conditions), and a Graph Component ($\mathcal{G}$) to capture the semantic ``Topology'' (e.g., interactions, hierarchies) for multi-hop reasoning. This hybrid architecture ensures both the flexibility required for complex reasoning and the rigor needed for numerical auditing.

Our approach addresses the full lifecycle of intelligent auditing. For construction, we propose an \textbf{Iterative Schema Refinement (ISR)} algorithm that dynamically evolves the hybrid schema to capture domain heterogeneity. This is implemented via a \textbf{Section-Aware Multi-Agent} framework that ensures every extracted fact is traceable to its source. For application, we introduce the \textbf{KB-grounded Chain of Verification (CoV)}. Instead of opaque generation, CoV decomposes audits into verifiable subtasks, executing transparent \textbf{Hybrid Queries} and utilizing a \textbf{Patient Profile-driven Evidence Selection Tree (P-EST)} to prune irrelevant rules. Crucially, the system is designed to explicitly flag \textbf{Information Gaps} when patient data is missing, prioritizing safety over hallucinated verdicts.

We evaluate PharmGraph-Auditor on a dataset of real-world inpatient prescriptions annotated by clinical experts. The results show that our framework consistently outperforms traditional rule-based CDSS, achieving a +13.4\% improvement in F1 scores. To provide deeper insights, we examine the system's ability to balance safety with efficiency. It significantly surpasses the recall of human experts while maintaining the high precision necessary to effectively mitigate pharmacist alert fatigue. While these results underscore its effectiveness in clinical settings, the hybrid architecture of PharmGraph-Auditor makes it adaptable to other complex domains requiring both rigorous constraint satisfaction and advanced semantic reasoning. Our code and data will be released upon publication.

\section{Methodology}
\label{sec:methodology}

To address the limitations of existing approaches, specifically their inability to simultaneously handle rigorous numerical auditing and complex semantic reasoning, we propose a formally grounded hybrid architecture. In this section, we establish the theoretical foundation of our data model based on the Virtual Knowledge Graph (VKG) paradigm, followed by the specific processes for its construction and application in prescription auditing.

\subsection{Theoretical Foundation: Hybrid VKG Model}
\label{sec:theoreticalmodel}

While vector databases dominate mainstream RAG architectures, they inherently lack the determinism required for clinical auditing, as semantic similarity cannot rigorously enforce exact numerical constraints. Consequently, a foundation in structured symbolic knowledge is essential. However, pharmaceutical knowledge possesses a dual nature, comprising both strict conditional rules and highly connected semantic concepts. A single structured data model proves insufficient: relational databases struggle with deep recursive reasoning, while pure graph databases lack efficient indexing mechanisms for complex range filtering. To resolve this dilemma, we frame our \textbf{Hybrid Pharmaceutical Knowledge Base (HPKB)} as a specialized implementation of the \textbf{Virtual Knowledge Graph (VKG)} paradigm \cite{xiao2019virtual, xiao2025llm4vkg}, adopting a \textbf{Hybrid Materialization} strategy.

Formally, we define the HPKB as a tuple $\mathcal{H} = \langle \mathcal{R}, \mathcal{G}, \phi \rangle$:

\begin{itemize}
    \item $\mathcal{R}$ is the \textbf{Relational Component} (Constraint Store), a set of relations $\{R_1, ..., R_n\}$ storing high-integrity atomic facts. It handles data requiring strict schema validation, such as dosage thresholds.
    \item $\mathcal{G}$ is the \textbf{Graph Component} (Topology Store), a labeled property graph $\mathcal{G} = (V, E)$ capturing the semantic topology of medical entities. It handles data requiring multi-hop reasoning.
    \item $\phi$ is the \textbf{Mapping Function}, $\phi: V \leftrightarrow \bigcup R_i$, a bijective function establishing explicit links between graph vertices and relational tuples, ensuring the system functions as a unified whole.
\end{itemize}

This hybrid architecture is not merely an engineering choice but a theoretical necessity derived from the algorithmic distinctness of prescription auditing tasks. To guide the schema design systematically, we propose the \textbf{Knowledge Stratification Framework} (Table \ref{tab:stratification}), which assigns data to $\mathcal{R}$ or $\mathcal{G}$ based on its logical nature.

\begin{table}[t]
  \caption{The Knowledge Stratification Framework}
  \label{tab:stratification}
  \centering
  \small
  \begin{tabular}{l p{2.8cm} p{2.8cm}}
    \toprule
    \textbf{Dimension} & \textbf{Relational ($\mathcal{R}$)} & \textbf{Graph ($\mathcal{G}$)} \\
    \midrule
    \textbf{Data Nature} & Atomic, Numerical, Conditional & Associative, Hierarchical, Transitive \\
    \addlinespace
    \textbf{Logic Type} & \textbf{Set Constraint Satisfaction} & \textbf{Topological Traversal} \\
    \addlinespace
    \textbf{Access Cost} & Index Scan: $O(\log N)$ & Index-free Adjacency: $O(1)^{\dagger}$ \\
    \addlinespace
    \textbf{Complexity} & Dependent on Dataset Size ($N$) & Independent of Dataset Size ($N$) \\
    \addlinespace
    \textbf{Typical Audit} & Dosage Checks, Contraindications & Interactions, Allergies, Duplicate Therapy \\
    \bottomrule
    \multicolumn{3}{l}{\footnotesize $^{\dagger}$ $O(1)$ denotes constant time per relationship traversal, irrespective of $|V|$.}
  \end{tabular}
\end{table}

\subsubsection{Set Constraint Satisfaction (Why Relational?)}
Audit tasks like \textit{Contraindication Checking} or \textit{Dosage Verification} are fundamentally \textbf{Set Constraint Satisfaction} problems. A clinical rule often manifests as a stack of boolean and range filters (e.g., \textit{Allow IF age $>$ 65 AND CrCl $<$ 30 AND hepatic\_impairment = `Severe'}).

We assign such data to $\mathcal{R}$ because Relational Database Management Systems (RDBMS) are mathematically optimized for Set Theory operations. They efficiently execute dynamic predicate logic through B-Tree indices, achieving $O(\log N)$ complexity for range lookups, where $N$ is the table cardinality. Modeling continuous numerical ranges in a Graph database would inherently require discretizing values into nodes or performing inefficient global property scans ($O(N)$), leading to unacceptable latency in real-time auditing.

\subsubsection{Topological Traversal (Why Graph?)}
Conversely, tasks like \textit{Interaction Screening} or \textit{Allergy Checking} are \textbf{Topological Traversal} problems involving path discovery and transitivity. For instance, detecting an allergy requires traversing a hierarchy: $\text{Patient} \xrightarrow{has} \text{Allergy} \xleftarrow{subclass\_of} \text{Concept} \xleftarrow{ingredient\_of} \text{Drug}$.

We assign such data to $\mathcal{G}$. The Graph model leverages \textbf{index-free adjacency}, where connected nodes physically point to each other in memory. This allows relationship traversal in constant time $O(1)$ per hop, completely independent of the total graph size ($|V|$ or $|E|$). In contrast, implementing deep recursion in $\mathcal{R}$ (e.g., via Recursive CTEs) necessitates iterative index lookups for each hop. Since every join operation incurs an $O(\log N)$ overhead, the cumulative cost for a path of depth $k$ scales as $O(k \cdot \log N)$. As the medical terminology hierarchy deepens and the dataset size $N$ grows, this logarithmic penalty accumulates, rendering relational recursion computationally brittle compared to the constant-time pointer dereferencing of graphs.

\subsection{Trustworthy HPKB Construction}
\label{sec:construction}

\begin{figure}[t]
	\centering
	\includegraphics[width=0.45\textwidth]{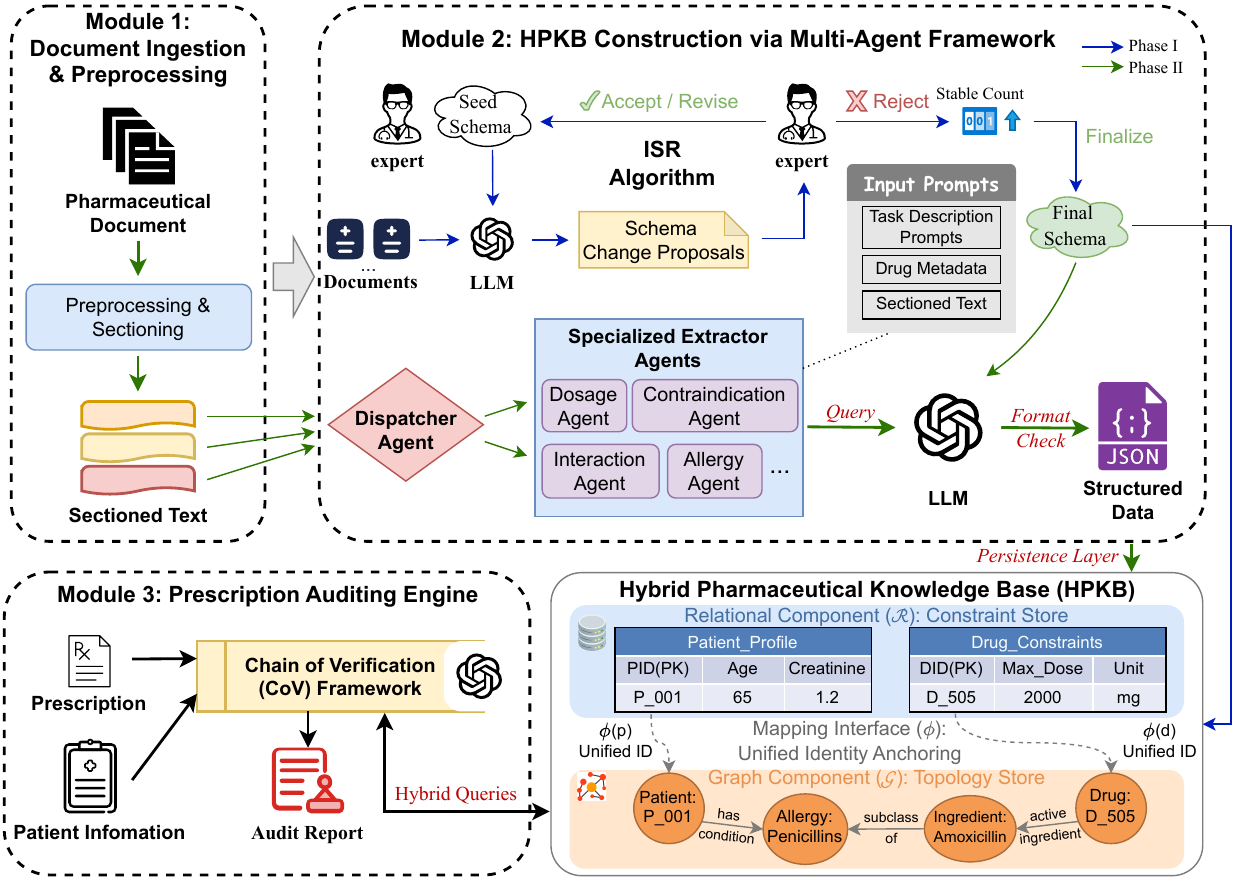}
	\caption{System architecture of PharmGraph-Auditor.}
	\label{fig:architecture}
\end{figure}

Guided by the theoretical model $\mathcal{H} = \langle \mathcal{R}, \mathcal{G}, \phi \rangle$, we implement a verifiable construction pipeline to populate the HPKB from unstructured pharmaceutical documents. As shown in Fig. \ref{fig:architecture}, the upper modules (Modules 1 \& 2) handle the pre-computation phase, preparing the knowledge base for the subsequent auditing inference.

\subsubsection{Phase I: Iterative Schema Refinement (ISR)}

Pre-defining a rigid, comprehensive schema for the pharmaceutical domain is impractical, given the sheer complexity and heterogeneity of medical knowledge. To address this, we propose the \textbf{Iterative Schema Refinement (ISR)} algorithm. This semi-automated, expert-supported process is designed to evolve a robust schema $\mathcal{S}_{final}$ that balances high information recall with structural compactness.

\paragraph{Stratified Sampling Strategy.}
To ensure the evolved schema generalizes across diverse medical contexts, the ISR process utilizes a stratified sampling strategy based on the ICD-10 classification (International Classification of Diseases, 10th Edition) \cite{icd_2019}. We select a representative corpus (e.g., 100 documents) evenly distributed across five major therapeutic areas with distinct structural complexities: \textit{Antineoplastic agents} (complex regimens), \textit{Anti-infectives} (contraindication-heavy), \textit{Cardiovascular agents} (interaction-heavy), \textit{Nervous system agents}, and \textit{Respiratory system agents}. This diversity is crucial for testing the schema's ability to handle heterogeneous data structures during the initialization phase.

\paragraph{Human-AI Synergy: Gap Detection and Abstraction.}

The ISR algorithm operates on a \textit{Propose-Verify-Solidify} loop that leverages the complementary strengths of Large Language Models (LLMs) and human experts. The process begins with a minimal ``seed schema''. For each document in the sample, the workflow proceeds as follows:

\begin{enumerate}
    \item \textbf{LLM as the Gap Detector:} The LLM agent compares the document text against the current schema. Beyond extracting fitting data, its primary role is to identify ``Schema Gaps''—valuable information (e.g., dosing prerequisites, infusion rates) that the current schema cannot represent. By drafting structured \textbf{Schema Change Proposals}, the LLM reduces the cognitive load on experts and prevents omissions common in manual construction.

    \item \textbf{Expert as the Architect:} Human experts review the proposals to enforce \textbf{Semantic Abstraction}. Our preliminary analysis suggests that LLMs, when unchecked, tend to suffer from ``Schema Fragmentation'' (e.g., proposing separate tables for \textit{RenalAdjustment} and \textit{HepaticAdjustment}). The expert mitigates this by elevating specific gaps into generalized structures. For instance, instead of accepting disparate fields for different organ functions, the expert defines a generic \texttt{Constraint} node. This abstract design ensures that the graph component $\mathcal{G}$ remains compact while uniformly representing renal, age-based, or weight-based restrictions.
\end{enumerate}

\paragraph{Decision Policy and Stabilization.}

The acceptance of schema proposals is governed by our Knowledge Stratification Framework. The expert classifies gaps into two categories to update the theoretical model $\mathcal{H} = \langle \mathcal{R}, \mathcal{G}, \phi \rangle$:

\begin{itemize}
    \item \textbf{Scenario A (Constraint Discovery):} If the agent encounters text defining a numerical boundary (e.g., Reduce dose by 50\% if CrCl $<$ 30 mL/min), the expert assigns it as a new attribute column in the Relational Component $\mathcal{R}$.
    \item \textbf{Scenario B (Topology Discovery):} If the agent identifies a connection between entities (e.g., Drug A is physically incompatible with Drug B), the expert assigns it as a new Edge Type (e.g., \texttt{has\_taboo}) in the Graph Component $\mathcal{G}$.
\end{itemize}

This iterative process continues until the schema exhibits ``rapid stabilization'', defined as the state where $N_{stable}$ consecutive documents pass without triggering valid schema change proposals. This ensures $\mathcal{S}_{final}$ achieving sufficient coverage of the domain's complexity.

\subsubsection{Phase II: Section-Aware Knowledge Population}

With the stable schema $\mathcal{S}_{final}$ established, we proceed to the full-scale population phase. The primary goal here is \textbf{verifiability}. We define a provenance function $\Pi(f) = (doc\_id, section, source\_text)$, requiring that every extracted fact $f$, whether a tuple in $\mathcal{R}$ or an edge in $\mathcal{G}$, must carry the exact source text from which it is derived.

\paragraph{1. Document Preprocessing and Sectioning}

The initial step is to convert heterogeneous source documents (typically PDFs) into a structured, machine-readable format. We leverage MinerU\footnote{https://github.com/opendatalab/MinerU}, an open-source toolkit that parses PDF documents into Markdown while preserving hierarchical structures like headers and tables. Given that pharmaceutical documents typically follow a standardized organization, this structured representation allows us to segment the text into distinct semantic blocks (e.g., \textit{Dosage and Administration}, \textit{Contraindications}, \textit{Drug Interactions}), enabling targeted, context-aware processing in subsequent steps.

\paragraph{2. The Section-Aware Multi-Agent Framework}
To overcome the context window limitations and attention drift often observed in monolithic LLMs, we introduce a Section-Aware Multi-Agent framework.
\begin{itemize}
    \item \textbf{The Dispatcher Agent:} Acting as the orchestrator, this agent analyzes the header of each text block and routes it to the appropriate specialist. For instance, a block labeled ``Drug Interactions'' is strictly routed to the Interaction Agent, minimizing noise.
    \item \textbf{Specialist Agents:} We deploy a suite of agents (e.g., \textit{Contraindication Agent}, \textit{Dosage Agent}, \textit{Interaction Agent}), each configured with a specialized prompt and a specific subset of $\mathcal{S}_{final}$ (derived from the ISR algorithm).
    \begin{itemize}
        \item The \textit{Dosage Agent} is prompted to extract structured tuples (e.g., $[age\_min, dose\_val]$) targeting the Relational Component $\mathcal{R}$.
        \item The \textit{Interaction Agent} is prompted to extract triples (e.g., $(Patient, has\_condition, Allergy)$) targeting the Graph Component $\mathcal{G}$.
    \end{itemize}
    Crucially, to ensure verifiability, each agent is strictly enforced to output the provenance metadata $\Pi(f)$ (the raw source text) alongside the extracted fact.
\end{itemize}

\paragraph{3. Hybrid Persistence and Mapping Layer}

The final component acts as a bridge between the agentic framework and the hybrid storage engine. It parses and validates the extracted JSON, routing data to either $\mathcal{R}$ or $\mathcal{G}$ according to the schema definitions. To operationalize the mapping $\phi$, we employ a Unified Identity Strategy, where a shared global identifier anchors entities across both storage modalities, implicitly ensuring data consistency and seamless cross-referencing.

\subsection{KB-Grounded Prescription Auditing}
\label{sec:auditing}

\begin{figure}[t]
	\centering
	\includegraphics[width=0.45\textwidth]{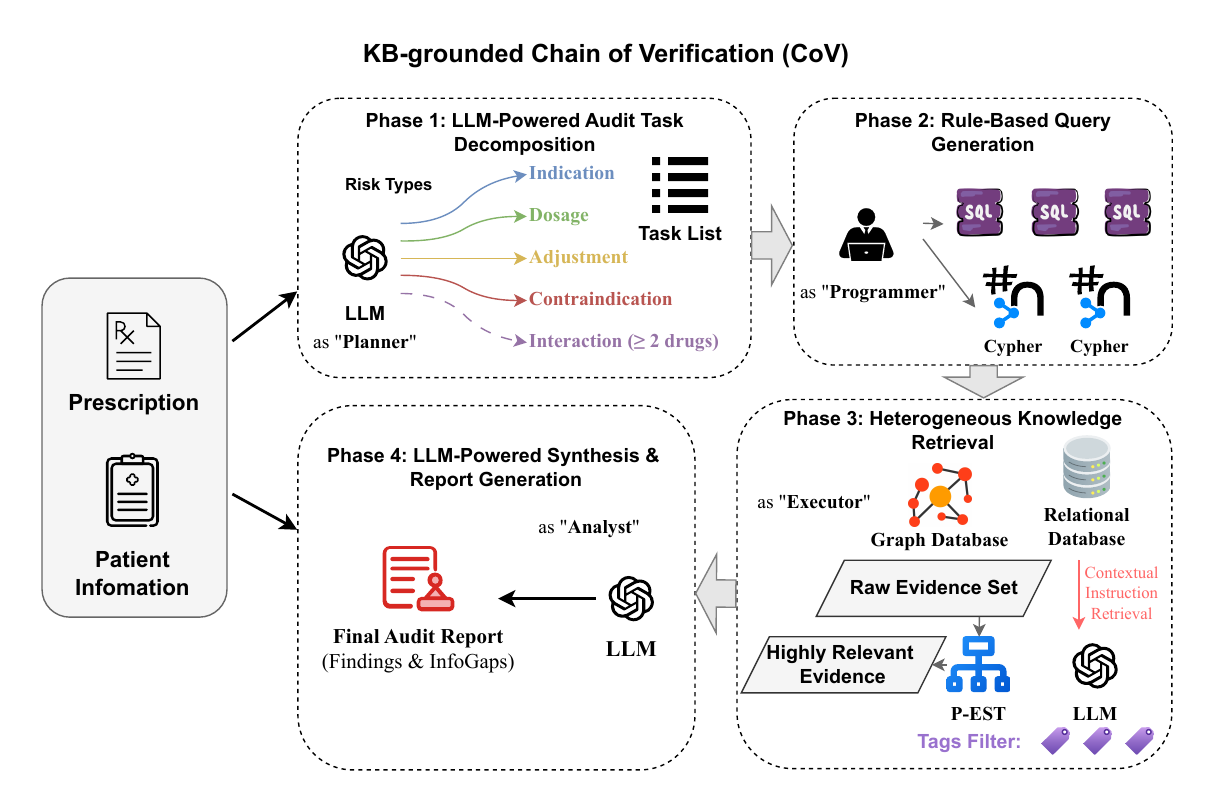}
	\caption{The KB-grounded Chain of Verification (CoV) framework.}
	\label{fig:cov}
\end{figure}

The ultimate goal of our system is to leverage the trustworthy HPKB to audit prescriptions. Formally, given a prescription instance $P = (\text{PatientInfo}, \text{DrugList})$, the task is to generate an audit report $(A, I)$, where $A$ contains evidence-grounded \textbf{Audit Findings} and $I$ identifies \textbf{Information Gaps} (missing patient data required for verification).

To achieve this in a manner that is safe, transparent, and robust against hallucination, we propose the \textbf{KB-grounded Chain of Verification (CoV)} framework. As illustrated in Fig. \ref{fig:cov} and detailed in Algorithm \ref{alg:cov}, CoV transforms the opaque ``black-box'' reasoning of LLMs into a transparent ``white-box'' pipeline comprising four distinct stages:

\begin{algorithm}[t]
\caption{KB-grounded Chain of Verification (CoV)}
\label{alg:cov}
\begin{algorithmic}[1]
    \State \textbf{Input:} Prescription $P$, Hybrid HPKB $\mathcal{K}$
    \State \textbf{Output:} Audit Report $(A, I)$
    \State $V_{plan} \gets \text{DecomposeTasks}(P)$ \Comment{Stage 1}
    \State $Evidence \gets \emptyset$
    
    \For{$task \in V_{plan}$}
        \State \Comment{Stage 2: Generate Hybrid Queries}
        \If{$task.type$ is Constraint}
            \State $Q \gets \text{GenerateSQL}(task)$
            \State $RawData \gets \mathcal{K}.\mathcal{R}.\text{execute}(Q)$
             \State \Comment{Stage 3: Curation via P-EST}
            \State $CuratedData \gets \text{P-EST}(P, RawData)$
        \Else
            \State $Q \gets \text{GenerateCypher}(task)$
            \State $CuratedData \gets \mathcal{K}.\mathcal{G}.\text{execute}(Q)$
        \EndIf
        \State $Evidence.add(task, CuratedData)$
    \EndFor
    
    \State \Comment{Stage 4: Synthesis}
    \State $(A, I) \gets \text{SynthesizeReport}(P, Evidence)$
    \State \Return $(A, I)$
\end{algorithmic}
\end{algorithm}

\subsubsection{Stage 1: LLM-driven Task Decomposition}
Instead of tasking an LLM with a monolithic instruction like ``check this prescription'', CoV first employs a specialized Decomposition Agent. This agent's sole responsibility is to decompose the high-level auditing goal into a \textbf{Verification Plan}. This plan is a structured list of specific, verifiable sub-tasks (e.g., \textit{dosage verification}, \textit{contraindication check}) tailored to the patient's profile and each prescribed drug.

\subsubsection{Stage 2: Hybrid Query Generation}
This stage is the operational core of our hybrid architecture. For each sub-task in the Verification Plan, a deterministic \textbf{Rule-based Query Engine} generates the precise database queries. Crucially, this engine selects the appropriate query language based on the task type, aligning with the theoretical stratification defined in Section \ref{sec:theoreticalmodel}:
\begin{itemize}
    \item \textbf{For Constraint Tasks (e.g., Dosage):} It generates SQL queries targeting the Relational Component $\mathcal{R}$ (e.g., ``SELECT * FROM DosageRules WHERE Drug=`Metformin' AND ...'').
    \item \textbf{For Topology Tasks (e.g., Allergy Analysis):} It generates Cypher queries targeting the Graph Component $\mathcal{G}$ to retrieve pharmacological hierarchies needed for reasoning. For instance, to screen for potential allergies, the query retrieves the drug's composition lineage (e.g., ``MATCH (d:Drug \{name: `Metformin'\})-[:HAS\_INGREDIENT]->(i:In- gredient )-[:BELONGS\_TO]->(c:Class) RETURN i, c'').
\end{itemize}
By relying on deterministic rules rather than LLM generation for query construction, we eliminate the risk of syntax errors or hallucinated database fields.

\subsubsection{Stage 3: Evidence Retrieval and Curation via P-EST}
Raw data retrieval is often insufficient. For instance, querying dosage rules for a generic drug might return dozens of rows covering various indications and populations. Feeding this ``noisy'' context to an LLM increases cognitive load and error rates.

\begin{figure}[t]
	\centering
	\includegraphics[width=0.45\textwidth]{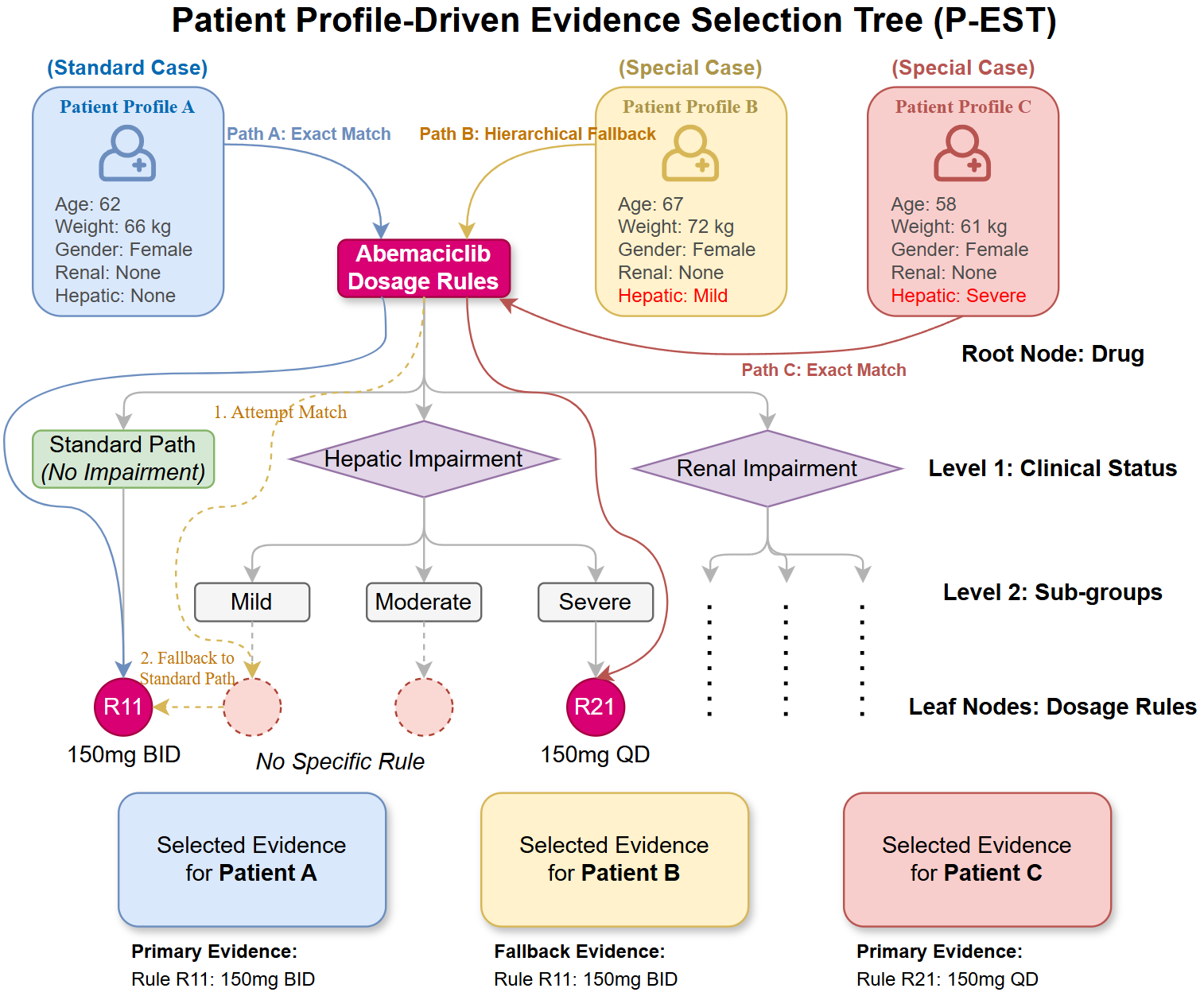}
	\caption{The Patient Profile-driven Evidence Selection Tree (P-EST) logic for pruning conflicting dosage rules.}
	\label{fig:p_est}
\end{figure}

To address this, we introduce the \textbf{Patient Profile-Driven Evidence Selection Tree (P-EST)} (Fig. \ref{fig:p_est}) for structured evidence curation. P-EST simulates clinical decision logic to prune irrelevant rules:
\begin{enumerate}
    \item \textbf{Exact Match Search:} It first attempts to find a rule that perfectly matches the patient's specific profile (e.g., ``Age 65, CrCl 25ml/min'').
    \item \textbf{Hierarchical Fallback:} If no exact match is found (common in real-world data), P-EST initiates a fallback search, moving up the decision tree to find the most specific applicable parent rule (e.g., ``Any Renal Impairment'') before defaulting to the standard adult dose.
\end{enumerate}
This ensures that the downstream LLM receives only the \textit{single most relevant} rule, maximizing precision.

\subsubsection{Stage 4: Evidence-Grounded Synthesis}
In the final stage, a Synthesis Agent receives the original prescription and the Curated Evidence Package to generate a structured \textbf{Audit Report}. 
To ensure safety, the agent integrates Uncertainty Handling into the synthesis process, explicitly identifying \textbf{Information Gaps ($I$)} whenever the evidence requirements exceed the available patient data. 
This is critical when retrieved evidence imposes \textit{conditional constraints}. For example, if the evidence states that \textit{``dosage must be reduced for patients with renal impairment''} but the patient profile in $P$ lacks renal function data, the agent flags this gap instead of hallucinating a verdict. 
The final output is a report where every finding is explicitly linked to the authoritative source text, ensuring full traceability.

\section{Experiments}

To comprehensively evaluate the effectiveness and robustness of \textit{PharmGraph-Auditor}, we design our experiments to answer two central research questions (RQs) that directly correspond to the core tasks defined in our problem formulation:

\quad \textit{RQ1: How effectively can PharmGraph-Auditor construct a high-fidelity HPKB?} This question assesses the core knowledge extraction pipeline, including the performance of our section-aware and multi-agent framework.

\quad \textit{RQ2: How accurately and safely does PharmGraph-Auditor perform evidence-grounded prescription auditing?} This question evaluates the application layer of our system, focusing on the performance of the KB-grounded CoV framework in identifying risks and handling uncertainty.

\subsection{Performance of HPKB Construction (RQ1)}
\label{sec:exp_rq1}

This section evaluates the system's ability to build the knowledge base. We focus on the quality of the knowledge population process, assuming a stable, expert-defined schema, to measure the performance of our extraction framework against established paradigms.

\subsubsection{Knowledge Population Quality}

\paragraph{Experimental Setup.}

\textbf{Evaluation Ground Truth:} To ensure a rigorous evaluation of the knowledge population performance, we constructed a gold-standard HPKB derived from 100 diverse pharmaceutical documents. The annotation process was conducted by a senior clinical pharmacist with over 10 years of experience in medical informatics and ontology construction. This expert meticulously reviewed the source documents and manually extracted all relevant entities, attribute values, and relations according to the final schema. This expert-curated dataset serves as the reliable ground truth for measuring the extraction fidelity of different systems. The statistics of this benchmark are detailed in Table \ref{tab:benchmark_stats}.

\begin{table}[tb]
\caption{Statistics of the Golden Standard HPKB}
\label{tab:benchmark_stats}
\centering
\resizebox{\columnwidth}{!}{%
    \begin{tabular}{lccc}
    \toprule
    \textbf{Metric} & \textbf{Relational Data} & \textbf{Graph Relations} & \textbf{Total} \\
    \midrule
    \# Documents & 100 & 100 & 100 \\
    \# Extracted Records & 2{,}951 & 923 & 3{,}874 \\
    Avg. \# Records / Doc & 29.51 & 9.23 & 38.74 \\
    \bottomrule
    \end{tabular}%
}
\end{table}

\begin{table*}[t]
  \caption{Performance on Knowledge Population Task}
  \label{tab:extraction_results}
  \centering
  \small 
  \renewcommand{\arraystretch}{1.2} 
  \begin{tabular}{llccc}
    \toprule
    \textbf{Method} & \textbf{Component} & \textbf{Precision} & \textbf{Recall} & \textbf{F1-score} \\
    \midrule
    \multirow{3}{*}{PharmGraph-Auditor (Ours, GPT-4o)} 
      & Relational & 0.7973 & 0.8243 & 0.8106 \\
      & Graph & 0.9565 & 0.9565 & 0.9565 \\
      & Overall (Micro) & 0.8260 & 0.8491 & 0.8374 \\
    \midrule
    \multirow{3}{*}{PharmGraph-Auditor (Ours, Deepseek-V3)} 
      & Relational & 0.7948 & 0.8378 & 0.8157 \\
      & Graph & 0.9782 & 0.9782 & 0.9782 \\
      & Overall (Micro) & 0.8235 & 0.8603 & 0.8415 \\
    \midrule
    \multirow{3}{*}{PharmGraph-Auditor (Ours, Qwen3-32B)} 
      & Relational & 0.8750 & 0.8513 & 0.8630 \\
      & Graph & 0.8750 & 0.9130 & 0.8936 \\
      & Overall (Micro) & 0.8750 & 0.8603 & 0.8676 \\
    \midrule
    \multirow{3}{*}{Zero-shot OpenIE (GraphRAG-style, GPT-4o)} 
      & Relational & 0.8409 & 0.5000 & 0.6271 \\
      & Graph & 0.8518 & 0.5000 & 0.6301 \\
      & Overall (Micro) & 0.8365 & 0.4860 & 0.6148 \\
    \midrule
    \multirow{3}{*}{One-shot Schema-guided (AutoKG-style, GPT-4o)} 
      & Relational & 0.7957 & 0.7635 & 0.7793 \\
      & Graph & 0.8125 & 0.8378 & 0.8297 \\
      & Overall (Micro) & 0.8023 & 0.7709 & 0.7863 \\
    \bottomrule
  \end{tabular}
\end{table*}

\textbf{Baselines:} We compare our section-aware, multi-agent framework (\textit{PharmGraph-Auditor}) against two strong baselines representing state-of-the-art approaches in knowledge extraction:
\begin{itemize}
    \item \textbf{Baseline 1: Zero-shot OpenIE (GraphRAG-style).} Simulating the indexing phase of GraphRAG \cite{edge2024local}, this baseline operates without pre-defined schema constraints. It utilizes a powerful LLM to process the full document and autonomously identify entities, extracting both structured attribute tuples (for relational records) and Subject-Verb-Object triples (for graph edges) in a bottom-up manner. This baseline represents the performance of unconstrained, general-purpose information extraction \textbf{covering both tabular and topological data}.
    \item \textbf{Baseline 2: One-shot Schema-guided Agent (AutoKG-style).} Adapted from the information extraction module of AutoKG \cite{zhu2024llms}, this baseline utilizes the full document context combined with a one-shot demonstration. To align with our hybrid data model, we extended its prompt to include not only candidate predicates for graph relations but also target fields for relational tables. It tests the efficacy of standard schema-guided prompting strategies against our fine-grained, section-aware approach.
\end{itemize}

\textbf{Metrics and Models:} We measure performance using standard Precision, Recall, and F1-score. To ensure a fair comparison, we calculate these metrics separately for the Relational and Graph components, as well as an overall micro-average. We implement our framework using three representative LLMs: GPT-4o, Deepseek-V3 \cite{liu2024deepseek}, and Qwen3-32B \cite{yang2025qwen3}, to assess the generalizability of our approach across different model scales and types.

\paragraph{Results.}

The quantitative results in Table \ref{tab:extraction_results} demonstrate that \textit{PharmGraph-Auditor} achieves state-of-the-art performance, consistently outperforming baselines across all metrics. Our framework maintains a robust balance between Recall and Precision, resulting in F1-scores exceeding 0.83 for all tested LLMs. This balance is critical: the high Recall ($>0.84$) ensures comprehensive coverage of potential medical risks, while the high Precision ($>0.82$) effectively suppresses hallucinations, validating the trustworthiness of the constructed knowledge base.

\textbf{Comparison with Baselines.} The baselines exhibit significant performance bottlenecks. The \textit{Zero-shot OpenIE} (GraphRAG-style), despite acceptable precision, suffers from a critically low Recall (0.4860), missing over half of the essential facts due to the lack of schema guidance. The \textit{One-shot Schema-guided} agent (AutoKG-style) improves Recall to 0.7709 but still falls short of our framework in both accuracy and completeness. These results confirm that processing complex documents in a single pass dilutes the model's attention (``Lost-in-the-Middle''), whereas our section-aware multi-agent approach ensures fine-grained and accurate extraction.

\begin{figure}[t]
	\centering
	\includegraphics[width=0.45\textwidth]{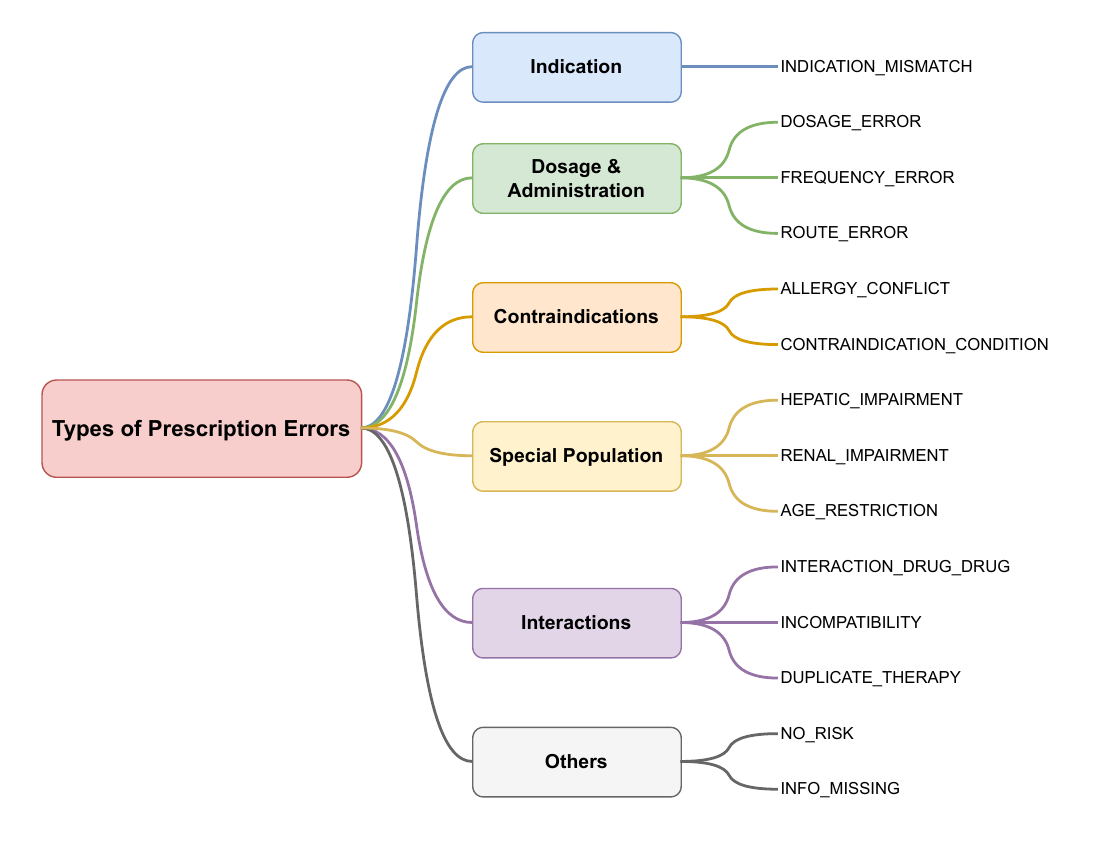}
	\caption{Types of prescription errors.}
	\label{fig:risk_type}
\end{figure}

\subsection{Performance of Prescription Auditing (RQ2)}
\label{sec:exp_rq2}

Having established the quality of our HPKB construction, we now evaluate the core application of our system: its ability to perform accurate, safe, and efficient prescription auditing in a real-world clinical setting.

\subsubsection{Experimental Setup}

\textbf{Real-world Dataset.} Unlike synthetic benchmarks, we conducted an experiment using 100 sets of authentic inpatient medical records and prescriptions from a real-world hospital. The dataset covers complex clinical scenarios across departments. We first defined a comprehensive taxonomy of prescription risks, categorized into five main types and their sub-types as illustrated in Fig. \ref{fig:risk_type}. Consequently, the auditing process focused on these five categories: \textit{Indications}, \textit{Dosage}, \textit{Contraindications}, \textit{Special Populations} and \textit{Interactions}, resulting in a total of 500 distinct audit points.

\textbf{Evaluation Baselines \& Process.} To rigorously assess the system, we employed a four-way comparative study:
\begin{enumerate}
    \item \textbf{Experience Review (Human Baseline):} Performed by a senior pharmacist with over 10 years of clinical experience, relying solely on professional memory and expertise.
    \item \textbf{Knowledge Review (Gold Standard):} The same senior pharmacist performed a second review, this time assisted by the retrieval results from our HPKB.
    \item \textbf{Proposed Method (CoV):} The audit performed automatically by our proposed HPKB-driven framework.
    \item \textbf{Rule Review (Traditional CDSS):} We compared our method against the hospital's legacy Clinical Decision Support System. Maintained for over 20 years, this pre-LLM system relies on rigid rules manually encoded by doctors and pharmacists, representing a labor-intensive construction process.
\end{enumerate}

The specific identified issues (37 out of 500) from the \textit{Knowledge Review} serve as the ground truth. This results in a correct prescription rate of 92.6\%, which aligns with real-world scenarios where baseline quality is generally high. However, given the zero-tolerance policy for errors in medical environments, an effective detection method remains essential to identify these infrequent but critical mistakes.

\subsubsection{Results and Analysis}

\begin{table}[t]
  \caption{Performance Comparison}
  \label{tab:auditing_comparison}
  \centering
  \resizebox{\columnwidth}{!}{%
    \begin{tabular}{lccc}
      \toprule
      \textbf{Method} & \textbf{Precision} & \textbf{Recall} & \textbf{F1-Score} \\
      \midrule
      Experience Review (Human) & 100.0\% & 45.9\% & 62.9\% \\
      Rule Review (CDSS) & 52.1\% & 67.6\% & 58.8\% \\
      \textbf{Proposed Method (Ours)} & \textbf{74.3\%} & \textbf{70.3\%} & \textbf{72.2\%} \\
      \bottomrule
    \end{tabular}%
  }
\end{table}

\textbf{Comparative Performance.}
Table \ref{tab:auditing_comparison} presents the performance of our AI method compared to the Human Baseline (\textit{Experience Review}) and the traditional rule-based CDSS. The results highlight distinct trade-offs inherent in each approach and validate the effectiveness of our proposed framework.

\textbf{Experience Review:} The human pharmacist achieved perfect Precision (100\%), indicating that seasoned experts rarely generate false alarms. However, this comes at the cost of safety coverage: the Recall was only 45.9\%. This result highlights the critical limitations of human memory and attention: more than half of the latent risks were missed when the pharmacist relied solely on experience. Notably, comparing this to the \textit{Gold Standard} (where the same pharmacist used our HPKB), the inclusion of retrieval capabilities boosted error detection by approximately 117\%. This significant gap confirms that high-fidelity retrieval is essential for comprehensive safety.
    
\textbf{Rule-based CDSS:} The traditional CDSS improved the Recall to 67.6\%, capturing a wider range of errors than the human baseline. However, it suffers from critically low Precision (52.1\%). Nearly half of the alerts generated by the rule engine were False Positives. In real-world clinical settings, such a high noise ratio significantly contributes to ``alert fatigue'', potentially causing pharmacists to ignore valid warnings.
    
\textbf{Proposed Method:} Our proposed method achieved the best balance between safety and efficiency. It surpassed the rule-based system in Recall (70.3\%), demonstrating superior sensitivity in detecting risks. Crucially, it achieved this while maintaining high Precision (74.3\%), significantly reducing the false-positive rate compared to the rule-based baseline.

Overall, these findings demonstrate that our approach serves as an effective assistant for human pharmacists. The knowledge retrieved by our system not only grounds the LLM's reasoning but also provides interpretable evidence for human verification. By achieving robust precision and recall through the CoV framework, our method acts as a reliable filter, minimizing both missed risks and false alarms, thereby significantly reducing the workload for clinical pharmacists.

\begin{figure}[t]
    \centering
    \includegraphics[width=0.48\textwidth]{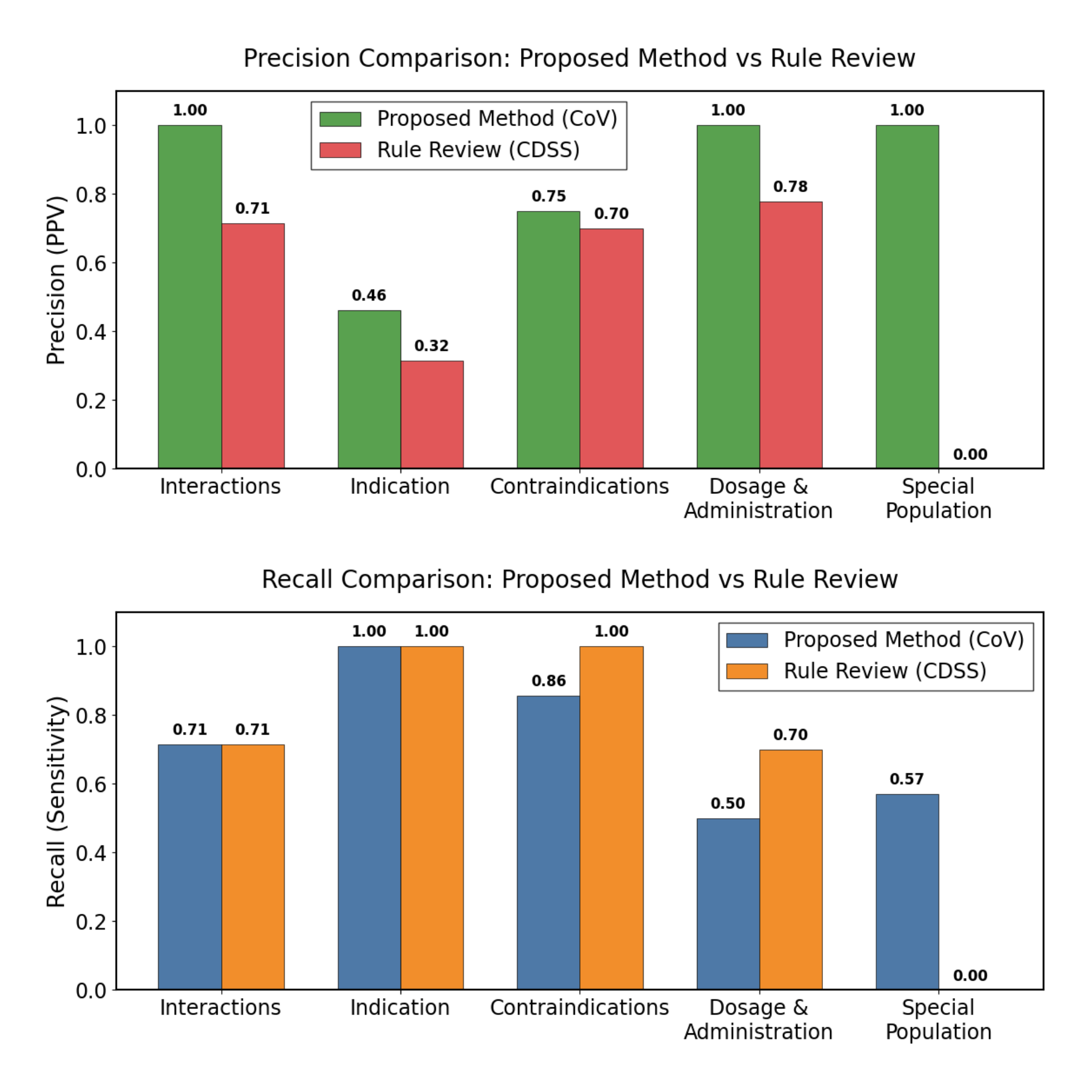}
    \caption{Fine-grained performance analysis by risk category. The top chart illustrates Precision, and the bottom chart illustrates Recall.}
    \label{fig:category_analysis}
\end{figure}

\textbf{Fine-grained Analysis by Risk Type.}
To understand the distinct behaviors of the two systems, we analyzed the results across specific error categories, as illustrated in Fig. \ref{fig:category_analysis}.

\textbf{Reasoning vs. Rigidity (Special Populations).}
The most profound divergence is observed in the \textit{Special Populations} category. The rule-based system exhibited a complete failure, failing to identify any of the true risks. In contrast, our proposed method successfully identified the majority of these issues. This performance gap highlights the fundamental limitation of rule-based logic: patient constraints (e.g., renal impairment status, geriatric frailty) are rarely stored in structured database fields. Instead, they are embedded within unstructured clinical notes and laboratory reports. Our method's semantic reasoning capability allows it to effectively infer these ``soft constraints'' from the context.

\begin{table}[t]
\caption{Ablation Study Results}
\label{tab:ablation_results}
\centering
\resizebox{\columnwidth}{!}{%
    \begin{tabular}{lcccc}
    \toprule
    \textbf{Setting} & \textbf{Precision} & \textbf{Recall} & \textbf{F1} & \textbf{Cost} \\
    \midrule
    \textbf{Proposed Method} & \textbf{0.7924} & \textbf{0.9504} & \textbf{0.8642} & \textbf{\$0.0225} \\
    w/o CoV & 0.5757 & 0.7645 & 0.6561 & \$0.0250 \\
    w/o CoV \& Knowledge & 0.3927 & 0.5233 & 0.4487 & \$0.0055 \\
    \bottomrule
    \end{tabular}%
}
\end{table}

\textbf{Precision and Alert Fatigue.}
Our method achieved superior Precision across \textit{all} risk categories compared to the rule-based baseline. Notably, we attained perfect or near-perfect precision in categories such as \textit{Interactions}, \textit{Dosage}, and \textit{Special Populations}, demonstrating exceptional reliability. The only outlier was the \textit{Indication} category; while our method still outperformed the baseline, the relatively lower precision scores for both systems highlight the inherent complexity of this specific task compared to the others. By generally maintaining such high precision, our approach directly mitigates the risk of pharmacist ``alert fatigue'', a critical factor in the practical adoption of safety systems.

\begin{figure}[t]
	\centering
	\includegraphics[width=0.5\textwidth]{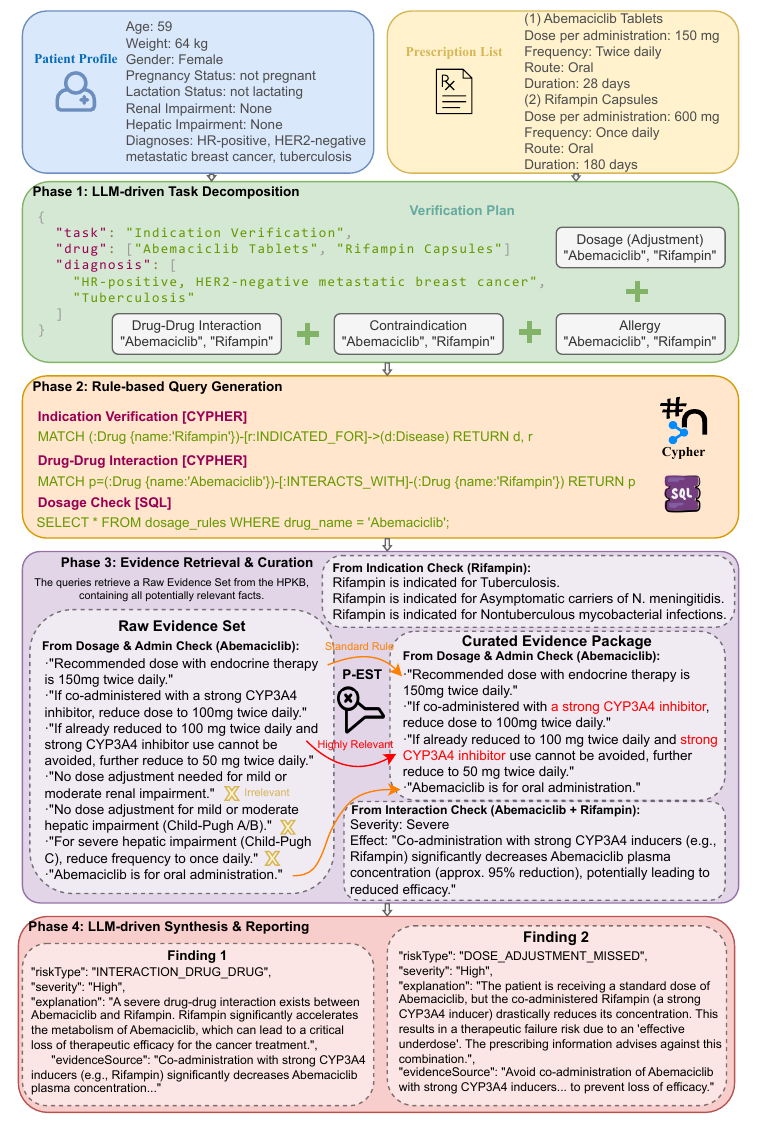}
	\caption{Case study.}
	\label{fig:case_study}
\end{figure}

\textbf{The Challenge of ``Clinical Context'' (Indications).}
Upon investigating the lower Precision (0.46) for \textit{Indications}, we identified that a significant portion of False Positives stemmed from a lack of ``Clinical Situational Awareness'', where the system strictly adheres to explicit pharmaceutical data while overlooking implicit clinical routines. A prime example is \textit{0.9\% Sodium Chloride} (Saline): in inpatient settings, it is routinely used as a solvent or for line flushing, but since these procedural utilities are not formally listed as ``therapeutic indications'' in package inserts, the LLM incorrectly flags them as mismatches. This finding highlights the necessity for future knowledge bases to be augmented with clinical procedural knowledge, enabling the differentiation between strict medical contraindications and accepted hospital practices.

\subsection{Ablation Study}
\label{sec:ablation}

To rigorously validate the contributions of the \textit{Chain of Verification (CoV)} framework and the external \textit{Knowledge Base}, we conducted an ablation study using a high-fidelity synthetic benchmark constructed via a ``Red Teaming'' approach.

\textbf{Dataset Construction.} We employed Gemini 2.5 Pro \cite{comanici2025gemini} to systematically generate over 1,000 prescription test cases. Specifically, the model was fed package inserts and prompted to synthesize erroneous prescriptions that deliberately violated the provided content. For each generated case, the model simultaneously identified the error type and cited the corresponding supporting evidence from the manual text. This generation was strictly guided by the comprehensive risk taxonomy illustrated in Fig. \ref{fig:risk_type}. To ensure data quality, expert pharmacists randomly sampled and reviewed the dataset, verifying that the prescriptions aligned with medical common sense and that the error logic strictly adhered to the prompts.

We compared our full method against two variants: (1) \textbf{w/o CoV Framework}, a standard Full-Text RAG approach where the LLM reads the entire drug package insert in a single pass; and (2) \textbf{w/o CoV \& w/o Knowledge}, a Zero-Shot setting relying solely on the LLM's internal weights.

As shown in Table \ref{tab:ablation_results}, removing external knowledge (\textit{w/o CoV \& Knowledge}) leads to a substantial decline in F1-score, confirming that internal parametric knowledge alone is insufficient for precise clinical auditing. While providing full documents (\textit{w/o CoV}) yields improvements over the zero-shot baseline, it still significantly underperforms our proposed method, primarily due to the noise inherent in processing lengthy unstructured texts. Notably, this standard RAG approach also incurs a higher cost than our proposed method, primarily due to excessive token consumption from processing full texts. Consequently, our full CoV method achieves the optimal balance of accuracy and efficiency by retrieving only precise, relevant facts.

\subsection{Case Study: Tracing the Chain of Verification}

To provide a concrete illustration of our system's end-to-end workflow, this section presents a case study of the KB-grounded Chain of Verification (CoV) framework in action. We trace a single, realistic prescription through the entire auditing pipeline, from initial task decomposition to the final, evidence-grounded report. This walkthrough, visualized in Figure \ref{fig:case_study}, demonstrates how the CoV's structured process ensures accuracy and safety by identifying a critical risk that less sophisticated methods might miss.

\textbf{Clinical Scenario:} We consider a complex but clinically realistic scenario involving a dose adjustment necessitated by a critical drug-drug interaction.

\begin{itemize}
    \item Patient Profile: A 59-year-old female patient is diagnosed with HR-positive, HER2-negative metastatic breast cancer and Tuberculosis. Her profile indicates no pre-existing renal or hepatic impairment.
    \item Prescription Order: The prescription includes Abemaciclib (150 mg twice daily) for her cancer. Critically, the patient is also being treated with Rifampin (600 mg once daily), a strong CYP3A4 inducer, for tuberculosis.
\end{itemize}

\section{Related Work}
\label{sec:relatedwork}

Prescription auditing, a critical component of medication safety, has evolved from manual pharmacist review to automated Clinical Decision Support Systems (CDSS). Traditional CDSS relies heavily on static, rule-based logic to detect errors such as drug-drug interactions (DDIs) and dosage violations \cite{bates1995incidence, kaushal2003effects, schiff2015computerised}. While effective in capturing explicit errors, these systems often suffer from high alert fatigue due to poor specificity and the inability to process unstructured clinical narratives \cite{nanji2014overrides}. To overcome these rigidity issues, researchers have increasingly integrated machine learning (ML) and deep learning techniques into auditing workflows. For instance, methods based on graph neural networks (GNNs) and molecular structure analysis have been proposed to predict DDIs with higher precision \cite{zitnik2018modeling, ryu2018deep, lin2020kgnn}. Furthermore, studies utilizing natural language processing (NLP) have demonstrated success in extracting medication entities from electronic health records (EHRs) to support more comprehensive automated checks \cite{meystre2008extracting, uzuner2010extracting, jensen2012mining}. Despite these advancements, most existing auditing systems remain siloed, handling either structured rules or unstructured text, but rarely synthesizing both for holistic risk assessment.

The advent of Large Language Models (LLMs) has introduced a transformative paradigm for clinical data processing. Foundational models fine-tuned on biomedical corpora, such as BioGPT \cite{luo2022biogpt}, PMC-LLaMA \cite{wu2024pmc}, and Med-PaLM 2 \cite{singhal2025toward}, have achieved expert performance in tasks from medical licensing exams to complex query answering \cite{nori2023capabilities, thirunavukarasu2023large}. To enhance comprehension, works like BALI \cite{sakhovskiy2025bali} propose augmenting these models by aligning textual representations with biomedical Knowledge Graphs. These capabilities suggest LLMs could function as comprehensive auditing agents. However, deploying LLMs in high-stakes clinical decision-making is hampered by hallucinations, where models generate plausible but incorrect assertions \cite{ji2023survey, zhang2025siren}. In prescription auditing, unsubstantiated outputs can lead to severe adverse drug events. Moreover, the opaque nature of LLM reasoning lacks the transparency required for clinical validation \cite{rudin2019stop, amann2020explainability}. To mitigate these risks, Retrieval-Augmented Generation (RAG) has emerged as a solution grounding model outputs in verified external knowledge \cite{lewis2020retrieval, guu2020retrieval}.

Recent research has advanced RAG by moving beyond unstructured text retrieval to leveraging structured knowledge sources, giving rise to GraphRAG and medical Knowledge Graph Question Answering (KGQA) systems. Frameworks like MedGraphRAG \cite{wu2025medical} and DoctorRAG \cite{lu2025doctorrag} utilize the topological structure of Knowledge Graphs (KGs) to enable multi-hop reasoning, allowing systems to trace relationships between symptoms, diagnoses, and treatments \cite{pan2024unifying, hu2024grag}. While these graph-centric approaches excel at semantic reasoning, they often struggle with the rigorous numerical constraint satisfaction problems inherent in prescription auditing, such as verifying renal function thresholds or weight-based dosage calculations \cite{hogan2021knowledge}. This limitation highlights the algorithmic distinctness between semantic traversal and set-based value filtering, suggesting that a single data model is insufficient for representing the full spectrum of pharmaceutical knowledge.

This data modeling challenge has revitalized interest in Hybrid Data Models and the Virtual Knowledge Graph (VKG) paradigm. VKG, historically known as Ontology-Based Data Access (OBDA), enables unified semantic querying over heterogeneous data sources without requiring physical migration to a graph format \cite{xiao2019virtual, calvanese2016ontop}. By mapping relational data to a conceptual graph layer, VKG frameworks allow systems to leverage the computational efficiency of SQL for numerical constraints while retaining the reasoning power of graph queries \cite{vrgoc2024millenniumdb, zhao2024hybrid}. Parallel advancements in Information Extraction (IE) and automated Knowledge Base Construction (KBC), such as AutoKG \cite{zhu2024llms} and schema-flexible extraction frameworks \cite{zhang2024extract, li2023type}, have made it feasible to populate such complex architectures from unstructured text. However, existing automated KBC pipelines often prioritize static schemas or single-model outputs, failing to address the dynamic need for stratifying knowledge into hybrid storage backends suitable for clinical auditing.

\section{Conclusion}

In this paper, we presented PharmGraph-Auditor, a hybrid framework designed to bridge the gap between LLM capabilities and the rigorous safety requirements of clinical prescription auditing. By formalizing the distinct nature of pharmaceutical knowledge through the relational modeling of numerical constraints and the graphical representation of semantic topology, our system ensures both precision and reasoning depth. We introduced the Iterative Schema Refinement (ISR) algorithm for trustworthy knowledge construction and the KB-grounded Chain of Verification (CoV) for transparent inference. Experimental evaluations confirm that our approach outperforms traditional CDSS by +13.4\% in F1 score, effectively mitigating pharmacist alert fatigue while significantly surpassing human experts in identifying latent risks.

Future work will focus on integrating Real-World Evidence (RWE) into the knowledge base to capture implicit clinical routines, bridging the gap between rigid pharmaceutical definitions and flexible hospital workflows. Ultimately, this work can present a new paradigm for building safe, traceable, and explainable AI systems for critical clinical decision support, shifting the LLM's role from an opaque generator to a verifiable, evidence-based reasoning engine.

\bibliographystyle{ACM-Reference-Format}
\bibliography{ref}


\begin{thebibliography}{45}


\ifx \showCODEN    \undefined \def \showCODEN     #1{\unskip}     \fi
\ifx \showISBNx    \undefined \def \showISBNx     #1{\unskip}     \fi
\ifx \showISBNxiii \undefined \def \showISBNxiii  #1{\unskip}     \fi
\ifx \showISSN     \undefined \def \showISSN      #1{\unskip}     \fi
\ifx \showLCCN     \undefined \def \showLCCN      #1{\unskip}     \fi
\ifx \shownote     \undefined \def \shownote      #1{#1}          \fi
\ifx \showarticletitle \undefined \def \showarticletitle #1{#1}   \fi
\ifx \showURL      \undefined \def \showURL       {\relax}        \fi
\providecommand\bibfield[2]{#2}
\providecommand\bibinfo[2]{#2}
\providecommand\natexlab[1]{#1}
\providecommand\showeprint[2][]{arXiv:#2}

\bibitem[Amann et~al\mbox{.}(2020)]%
        {amann2020explainability}
\bibfield{author}{\bibinfo{person}{Julia Amann}, \bibinfo{person}{Alessandro Blasimme}, \bibinfo{person}{Effy Vayena}, \bibinfo{person}{Dietmar Frey}, \bibinfo{person}{Vince~I Madai}, {and} \bibinfo{person}{Precise4Q Consortium}.} \bibinfo{year}{2020}\natexlab{}.
\newblock \showarticletitle{Explainability for artificial intelligence in healthcare: A multidisciplinary perspective}.
\newblock \bibinfo{journal}{\emph{BMC Medical Informatics and Decision Making}} \bibinfo{volume}{20}, \bibinfo{number}{1} (\bibinfo{year}{2020}), \bibinfo{pages}{310}.
\newblock


\bibitem[Aspden and Aspden(2007)]%
        {aspden2007preventing}
\bibfield{author}{\bibinfo{person}{Philip Aspden} {and} \bibinfo{person}{Philip Aspden}.} \bibinfo{year}{2007}\natexlab{}.
\newblock \bibinfo{booktitle}{\emph{Preventing medication errors}}. Vol.~\bibinfo{volume}{8}.
\newblock \bibinfo{publisher}{National Academies Press Washington, DC}.
\newblock


\bibitem[Barone et~al\mbox{.}(2025)]%
        {barone2025combining}
\bibfield{author}{\bibinfo{person}{Mariano Barone}, \bibinfo{person}{Antonio Romano}, \bibinfo{person}{Giuseppe Riccio}, \bibinfo{person}{Marco Postiglione}, {and} \bibinfo{person}{Vincenzo Moscato}.} \bibinfo{year}{2025}\natexlab{}.
\newblock \showarticletitle{Combining evidence and reasoning for biomedical fact-checking}. In \bibinfo{booktitle}{\emph{Proceedings of the 48th International ACM SIGIR Conference on Research and Development in Information Retrieval}}. \bibinfo{pages}{1087--1097}.
\newblock


\bibitem[Bates et~al\mbox{.}(1995)]%
        {bates1995incidence}
\bibfield{author}{\bibinfo{person}{David~W Bates}, \bibinfo{person}{David~J Cullen}, \bibinfo{person}{Nan Laird}, \bibinfo{person}{Laura~A Petersen}, \bibinfo{person}{Stephen~D Small}, \bibinfo{person}{Deborah Servi}, \bibinfo{person}{Glenn Laffel}, \bibinfo{person}{Bobbie~J Sweitzer}, \bibinfo{person}{Brian~F Shea}, \bibinfo{person}{Robert Hallisey}, {et~al\mbox{.}}} \bibinfo{year}{1995}\natexlab{}.
\newblock \showarticletitle{Incidence of adverse drug events and potential adverse drug events: Implications for prevention}.
\newblock \bibinfo{journal}{\emph{Jama}} \bibinfo{volume}{274}, \bibinfo{number}{1} (\bibinfo{year}{1995}), \bibinfo{pages}{29--34}.
\newblock


\bibitem[Calvanese et~al\mbox{.}(2016)]%
        {calvanese2016ontop}
\bibfield{author}{\bibinfo{person}{Diego Calvanese}, \bibinfo{person}{Benjamin Cogrel}, \bibinfo{person}{Sarah Komla-Ebri}, \bibinfo{person}{Roman Kontchakov}, \bibinfo{person}{Davide Lanti}, \bibinfo{person}{Martin Rezk}, \bibinfo{person}{Mariano Rodriguez-Muro}, {and} \bibinfo{person}{Guohui Xiao}.} \bibinfo{year}{2016}\natexlab{}.
\newblock \showarticletitle{Ontop: Answering SPARQL queries over relational databases}.
\newblock \bibinfo{journal}{\emph{Semantic Web}} \bibinfo{volume}{8}, \bibinfo{number}{3} (\bibinfo{year}{2016}), \bibinfo{pages}{471--487}.
\newblock


\bibitem[Comanici et~al\mbox{.}(2025)]%
        {comanici2025gemini}
\bibfield{author}{\bibinfo{person}{Gheorghe Comanici}, \bibinfo{person}{Eric Bieber}, \bibinfo{person}{Mike Schaekermann}, \bibinfo{person}{Ice Pasupat}, \bibinfo{person}{Noveen Sachdeva}, \bibinfo{person}{Inderjit Dhillon}, \bibinfo{person}{Marcel Blistein}, \bibinfo{person}{Ori Ram}, \bibinfo{person}{Dan Zhang}, \bibinfo{person}{Evan Rosen}, {et~al\mbox{.}}} \bibinfo{year}{2025}\natexlab{}.
\newblock \showarticletitle{Gemini 2.5: Pushing the frontier with advanced reasoning, multimodality, long context, and next generation agentic capabilities}.
\newblock \bibinfo{journal}{\emph{arXiv preprint arXiv:2507.06261}} (\bibinfo{year}{2025}).
\newblock


\bibitem[Edge et~al\mbox{.}(2024)]%
        {edge2024local}
\bibfield{author}{\bibinfo{person}{Darren Edge}, \bibinfo{person}{Ha Trinh}, \bibinfo{person}{Newman Cheng}, \bibinfo{person}{Joshua Bradley}, \bibinfo{person}{Alex Chao}, \bibinfo{person}{Apurva Mody}, \bibinfo{person}{Steven Truitt}, \bibinfo{person}{Dasha Metropolitansky}, \bibinfo{person}{Robert~Osazuwa Ness}, {and} \bibinfo{person}{Jonathan Larson}.} \bibinfo{year}{2024}\natexlab{}.
\newblock \showarticletitle{From local to global: A graph {RAG} approach to query-focused summarization}.
\newblock \bibinfo{journal}{\emph{arXiv preprint arXiv:2404.16130}} (\bibinfo{year}{2024}).
\newblock


\bibitem[Gorbach et~al\mbox{.}(2015)]%
        {gorbach2015frequency}
\bibfield{author}{\bibinfo{person}{Christy Gorbach}, \bibinfo{person}{Linda Blanton}, \bibinfo{person}{Beverly~A Lukawski}, \bibinfo{person}{Alex~C Varkey}, \bibinfo{person}{E~Paige Pitman}, {and} \bibinfo{person}{Kevin~W Garey}.} \bibinfo{year}{2015}\natexlab{}.
\newblock \showarticletitle{Frequency of and risk factors for medication errors by pharmacists during order verification in a tertiary care medical center}.
\newblock \bibinfo{journal}{\emph{American Journal of Health-System Pharmacy}} \bibinfo{volume}{72}, \bibinfo{number}{17} (\bibinfo{year}{2015}), \bibinfo{pages}{1471--1474}.
\newblock


\bibitem[Guu et~al\mbox{.}(2020)]%
        {guu2020retrieval}
\bibfield{author}{\bibinfo{person}{Kelvin Guu}, \bibinfo{person}{Kenton Lee}, \bibinfo{person}{Zora Tung}, \bibinfo{person}{Panupong Pasupat}, {and} \bibinfo{person}{Mingwei Chang}.} \bibinfo{year}{2020}\natexlab{}.
\newblock \showarticletitle{Retrieval augmented language model pre-training}. In \bibinfo{booktitle}{\emph{International Conference on Machine Learning}}. PMLR, \bibinfo{pages}{3929--3938}.
\newblock


\bibitem[Hogan et~al\mbox{.}(2021)]%
        {hogan2021knowledge}
\bibfield{author}{\bibinfo{person}{Aidan Hogan}, \bibinfo{person}{Eva Blomqvist}, \bibinfo{person}{Michael Cochez}, \bibinfo{person}{Claudia d’Amato}, \bibinfo{person}{Gerard~De Melo}, \bibinfo{person}{Claudio Gutierrez}, \bibinfo{person}{Sabrina Kirrane}, \bibinfo{person}{Jos{\'e} Emilio~Labra Gayo}, \bibinfo{person}{Roberto Navigli}, \bibinfo{person}{Sebastian Neumaier}, {et~al\mbox{.}}} \bibinfo{year}{2021}\natexlab{}.
\newblock \showarticletitle{Knowledge graphs}.
\newblock \bibinfo{journal}{\emph{ACM Computing Surveys (Csur)}} \bibinfo{volume}{54}, \bibinfo{number}{4} (\bibinfo{year}{2021}), \bibinfo{pages}{1--37}.
\newblock


\bibitem[Hu et~al\mbox{.}(2024)]%
        {hu2024grag}
\bibfield{author}{\bibinfo{person}{Yuntong Hu}, \bibinfo{person}{Zhihan Lei}, \bibinfo{person}{Zheng Zhang}, \bibinfo{person}{Bo Pan}, \bibinfo{person}{Chen Ling}, {and} \bibinfo{person}{Liang Zhao}.} \bibinfo{year}{2024}\natexlab{}.
\newblock \showarticletitle{{GRAG}: Graph Retrieval-Augmented Generation}.
\newblock \bibinfo{journal}{\emph{arXiv preprint arXiv:2405.16506}} (\bibinfo{year}{2024}).
\newblock


\bibitem[Jensen et~al\mbox{.}(2012)]%
        {jensen2012mining}
\bibfield{author}{\bibinfo{person}{Peter~B Jensen}, \bibinfo{person}{Lars~J Jensen}, {and} \bibinfo{person}{S{\o}ren Brunak}.} \bibinfo{year}{2012}\natexlab{}.
\newblock \showarticletitle{Mining electronic health records: towards better research applications and clinical care}.
\newblock \bibinfo{journal}{\emph{Nature Reviews Genetics}} \bibinfo{volume}{13}, \bibinfo{number}{6} (\bibinfo{year}{2012}), \bibinfo{pages}{395--405}.
\newblock


\bibitem[Ji et~al\mbox{.}(2023)]%
        {ji2023survey}
\bibfield{author}{\bibinfo{person}{Ziwei Ji}, \bibinfo{person}{Nayeon Lee}, \bibinfo{person}{Rita Frieske}, \bibinfo{person}{Tiezheng Yu}, \bibinfo{person}{Dan Su}, \bibinfo{person}{Yan Xu}, \bibinfo{person}{Etsuko Ishii}, \bibinfo{person}{Ye~Jin Bang}, \bibinfo{person}{Andrea Madotto}, {and} \bibinfo{person}{Pascale Fung}.} \bibinfo{year}{2023}\natexlab{}.
\newblock \showarticletitle{Survey of hallucination in natural language generation}.
\newblock \bibinfo{journal}{\emph{Comput. Surveys}} \bibinfo{volume}{55}, \bibinfo{number}{12} (\bibinfo{year}{2023}), \bibinfo{pages}{1--38}.
\newblock


\bibitem[Kaushal et~al\mbox{.}(2003)]%
        {kaushal2003effects}
\bibfield{author}{\bibinfo{person}{Rainu Kaushal}, \bibinfo{person}{Kaveh~G Shojania}, {and} \bibinfo{person}{David~W Bates}.} \bibinfo{year}{2003}\natexlab{}.
\newblock \showarticletitle{Effects of computerized physician order entry and clinical decision support systems on medication safety: a systematic review}.
\newblock \bibinfo{journal}{\emph{Archives of Internal Medicine}} \bibinfo{volume}{163}, \bibinfo{number}{12} (\bibinfo{year}{2003}), \bibinfo{pages}{1409--1416}.
\newblock


\bibitem[Lewis et~al\mbox{.}(2020)]%
        {lewis2020retrieval}
\bibfield{author}{\bibinfo{person}{Patrick Lewis}, \bibinfo{person}{Ethan Perez}, \bibinfo{person}{Aleksandra Piktus}, \bibinfo{person}{Fabio Petroni}, \bibinfo{person}{Vladimir Karpukhin}, \bibinfo{person}{Naman Goyal}, \bibinfo{person}{Heinrich K{\"u}ttler}, \bibinfo{person}{Mike Lewis}, \bibinfo{person}{Wen-tau Yih}, \bibinfo{person}{Tim Rockt{\"a}schel}, {et~al\mbox{.}}} \bibinfo{year}{2020}\natexlab{}.
\newblock \showarticletitle{Retrieval-augmented generation for knowledge-intensive {NLP} tasks}.
\newblock \bibinfo{journal}{\emph{Advances in Neural Information Processing Systems}}  \bibinfo{volume}{33} (\bibinfo{year}{2020}), \bibinfo{pages}{9459--9474}.
\newblock


\bibitem[Li et~al\mbox{.}(2023)]%
        {li2023type}
\bibfield{author}{\bibinfo{person}{Yongqi Li}, \bibinfo{person}{Yu Yu}, {and} \bibinfo{person}{Tieyun Qian}.} \bibinfo{year}{2023}\natexlab{}.
\newblock \showarticletitle{Type-aware decomposed framework for few-shot named entity recognition}. In \bibinfo{booktitle}{\emph{Findings of the Association for Computational Linguistics: EMNLP 2023}}. \bibinfo{pages}{8911--8927}.
\newblock


\bibitem[Lin et~al\mbox{.}(2020)]%
        {lin2020kgnn}
\bibfield{author}{\bibinfo{person}{Xuan Lin}, \bibinfo{person}{Zhe Quan}, \bibinfo{person}{Zhi-Jie Wang}, \bibinfo{person}{Tengfei Ma}, {and} \bibinfo{person}{Xiangxiang Zeng}.} \bibinfo{year}{2020}\natexlab{}.
\newblock \showarticletitle{KGNN: Knowledge graph neural network for drug-drug interaction prediction}. In \bibinfo{booktitle}{\emph{IJCAI}}, Vol.~\bibinfo{volume}{380}. \bibinfo{pages}{2739--2745}.
\newblock


\bibitem[Liu et~al\mbox{.}(2024)]%
        {liu2024deepseek}
\bibfield{author}{\bibinfo{person}{Aixin Liu}, \bibinfo{person}{Bei Feng}, \bibinfo{person}{Bing Xue}, \bibinfo{person}{Bingxuan Wang}, \bibinfo{person}{Bochao Wu}, \bibinfo{person}{Chengda Lu}, \bibinfo{person}{Chenggang Zhao}, \bibinfo{person}{Chengqi Deng}, \bibinfo{person}{Chenyu Zhang}, \bibinfo{person}{Chong Ruan}, {et~al\mbox{.}}} \bibinfo{year}{2024}\natexlab{}.
\newblock \showarticletitle{Deepseek-v3 technical report}.
\newblock \bibinfo{journal}{\emph{arXiv preprint arXiv:2412.19437}} (\bibinfo{year}{2024}).
\newblock


\bibitem[Lu et~al\mbox{.}(2025)]%
        {lu2025doctorrag}
\bibfield{author}{\bibinfo{person}{Yuxing Lu}, \bibinfo{person}{Gecheng Fu}, \bibinfo{person}{Wei Wu}, \bibinfo{person}{Xukai Zhao}, \bibinfo{person}{Goi~Sin Yee}, {and} \bibinfo{person}{Jinzhuo Wang}.} \bibinfo{year}{2025}\natexlab{}.
\newblock \showarticletitle{Towards doctor-like reasoning: Medical {RAG} fusing knowledge with patient analogy through textual gradients}. In \bibinfo{booktitle}{\emph{39th Annual Conference on Neural Information Processing Systems}}.
\newblock


\bibitem[Luo et~al\mbox{.}(2022)]%
        {luo2022biogpt}
\bibfield{author}{\bibinfo{person}{Renqian Luo}, \bibinfo{person}{Liai Sun}, \bibinfo{person}{Yingce Xia}, \bibinfo{person}{Tao Qin}, \bibinfo{person}{Sheng Zhang}, \bibinfo{person}{Hoifung Poon}, {and} \bibinfo{person}{Tie-Yan Liu}.} \bibinfo{year}{2022}\natexlab{}.
\newblock \showarticletitle{BioGPT: Generative pre-trained transformer for biomedical text generation and mining}.
\newblock \bibinfo{journal}{\emph{Briefings in Bioinformatics}} \bibinfo{volume}{23}, \bibinfo{number}{6} (\bibinfo{year}{2022}), \bibinfo{pages}{bbac409}.
\newblock


\bibitem[Meystre et~al\mbox{.}(2008)]%
        {meystre2008extracting}
\bibfield{author}{\bibinfo{person}{St{\'e}phane~M Meystre}, \bibinfo{person}{Guergana~K Savova}, \bibinfo{person}{Karin~C Kipper-Schuler}, {and} \bibinfo{person}{John~F Hurdle}.} \bibinfo{year}{2008}\natexlab{}.
\newblock \showarticletitle{Extracting information from textual documents in the electronic health record: A review of recent research}.
\newblock \bibinfo{journal}{\emph{Yearbook of Medical Informatics}} \bibinfo{volume}{17}, \bibinfo{number}{01} (\bibinfo{year}{2008}), \bibinfo{pages}{128--144}.
\newblock


\bibitem[Nanji et~al\mbox{.}(2014)]%
        {nanji2014overrides}
\bibfield{author}{\bibinfo{person}{Karen~C Nanji}, \bibinfo{person}{Sarah~P Slight}, \bibinfo{person}{Diane~L Seger}, \bibinfo{person}{Insook Cho}, \bibinfo{person}{Julie~M Fiskio}, \bibinfo{person}{Lisa~M Redden}, \bibinfo{person}{Lynn~A Volk}, {and} \bibinfo{person}{David~W Bates}.} \bibinfo{year}{2014}\natexlab{}.
\newblock \showarticletitle{Overrides of medication-related clinical decision support alerts in outpatients}.
\newblock \bibinfo{journal}{\emph{Journal of the American Medical Informatics Association}} \bibinfo{volume}{21}, \bibinfo{number}{3} (\bibinfo{year}{2014}), \bibinfo{pages}{487--491}.
\newblock


\bibitem[Nori et~al\mbox{.}(2023)]%
        {nori2023capabilities}
\bibfield{author}{\bibinfo{person}{Harsha Nori}, \bibinfo{person}{Nicholas King}, \bibinfo{person}{Scott~Mayer McKinney}, \bibinfo{person}{Dean Carignan}, {and} \bibinfo{person}{Eric Horvitz}.} \bibinfo{year}{2023}\natexlab{}.
\newblock \showarticletitle{Capabilities of {GPT}-4 on medical challenge problems}.
\newblock \bibinfo{journal}{\emph{arXiv preprint arXiv:2303.13375}} (\bibinfo{year}{2023}).
\newblock


\bibitem[Pais et~al\mbox{.}(2024)]%
        {pais2024large}
\bibfield{author}{\bibinfo{person}{Cristobal Pais}, \bibinfo{person}{Jianfeng Liu}, \bibinfo{person}{Robert Voigt}, \bibinfo{person}{Vin Gupta}, \bibinfo{person}{Elizabeth Wade}, {and} \bibinfo{person}{Mohsen Bayati}.} \bibinfo{year}{2024}\natexlab{}.
\newblock \showarticletitle{Large language models for preventing medication direction errors in online pharmacies}.
\newblock \bibinfo{journal}{\emph{Nature Medicine}} \bibinfo{volume}{30}, \bibinfo{number}{6} (\bibinfo{year}{2024}), \bibinfo{pages}{1574--1582}.
\newblock


\bibitem[Pan et~al\mbox{.}(2024)]%
        {pan2024unifying}
\bibfield{author}{\bibinfo{person}{Shirui Pan}, \bibinfo{person}{Linhao Luo}, \bibinfo{person}{Yufei Wang}, \bibinfo{person}{Chen Chen}, \bibinfo{person}{Jiapu Wang}, {and} \bibinfo{person}{Xindong Wu}.} \bibinfo{year}{2024}\natexlab{}.
\newblock \showarticletitle{Unifying large language models and knowledge graphs: A roadmap}.
\newblock \bibinfo{journal}{\emph{IEEE Transactions on Knowledge and Data Engineering}} \bibinfo{volume}{36}, \bibinfo{number}{7} (\bibinfo{year}{2024}), \bibinfo{pages}{3580--3599}.
\newblock


\bibitem[Rudin(2019)]%
        {rudin2019stop}
\bibfield{author}{\bibinfo{person}{Cynthia Rudin}.} \bibinfo{year}{2019}\natexlab{}.
\newblock \showarticletitle{Stop explaining black box machine learning models for high stakes decisions and use interpretable models instead}.
\newblock \bibinfo{journal}{\emph{Nature Machine Intelligence}} \bibinfo{volume}{1}, \bibinfo{number}{5} (\bibinfo{year}{2019}), \bibinfo{pages}{206--215}.
\newblock


\bibitem[Ryu et~al\mbox{.}(2018)]%
        {ryu2018deep}
\bibfield{author}{\bibinfo{person}{Jae~Yong Ryu}, \bibinfo{person}{Hyun~Uk Kim}, {and} \bibinfo{person}{Sang~Yup Lee}.} \bibinfo{year}{2018}\natexlab{}.
\newblock \showarticletitle{Deep learning improves prediction of drug--drug and drug--food interactions}.
\newblock \bibinfo{journal}{\emph{Proceedings of the National Academy of Sciences}} \bibinfo{volume}{115}, \bibinfo{number}{18} (\bibinfo{year}{2018}), \bibinfo{pages}{E4304--E4311}.
\newblock


\bibitem[Sakhovskiy and Tutubalina(2025)]%
        {sakhovskiy2025bali}
\bibfield{author}{\bibinfo{person}{Andrey Sakhovskiy} {and} \bibinfo{person}{Elena Tutubalina}.} \bibinfo{year}{2025}\natexlab{}.
\newblock \showarticletitle{{BALI}: Enhancing biomedical language representations through knowledge graph and language model alignment}. In \bibinfo{booktitle}{\emph{Proceedings of the 48th International ACM SIGIR Conference on Research and Development in Information Retrieval}}. \bibinfo{pages}{1152--1164}.
\newblock


\bibitem[Schiff et~al\mbox{.}(2015)]%
        {schiff2015computerised}
\bibfield{author}{\bibinfo{person}{Gordon~D Schiff}, \bibinfo{person}{MG Amato}, \bibinfo{person}{T Eguale}, \bibinfo{person}{JJ Boehne}, \bibinfo{person}{A Wright}, \bibinfo{person}{R Koppel}, \bibinfo{person}{AH Rashidee}, \bibinfo{person}{RB Elson}, \bibinfo{person}{DL Whitney}, \bibinfo{person}{TT Thach}, {et~al\mbox{.}}} \bibinfo{year}{2015}\natexlab{}.
\newblock \showarticletitle{Computerised physician order entry-related medication errors: Analysis of reported errors and vulnerability testing of current systems}.
\newblock \bibinfo{journal}{\emph{BMJ Quality \& Safety}} \bibinfo{volume}{24}, \bibinfo{number}{4} (\bibinfo{year}{2015}), \bibinfo{pages}{264--271}.
\newblock


\bibitem[Singhal et~al\mbox{.}(2025)]%
        {singhal2025toward}
\bibfield{author}{\bibinfo{person}{Karan Singhal}, \bibinfo{person}{Tao Tu}, \bibinfo{person}{Juraj Gottweis}, \bibinfo{person}{Rory Sayres}, \bibinfo{person}{Ellery Wulczyn}, \bibinfo{person}{Mohamed Amin}, \bibinfo{person}{Le Hou}, \bibinfo{person}{Kevin Clark}, \bibinfo{person}{Stephen~R Pfohl}, \bibinfo{person}{Heather Cole-Lewis}, {et~al\mbox{.}}} \bibinfo{year}{2025}\natexlab{}.
\newblock \showarticletitle{Toward expert-level medical question answering with large language models}.
\newblock \bibinfo{journal}{\emph{Nature Medicine}} \bibinfo{volume}{31}, \bibinfo{number}{3} (\bibinfo{year}{2025}), \bibinfo{pages}{943--950}.
\newblock


\bibitem[Sridharan and Sivaramakrishnan(2024)]%
        {sridharan2024unlocking}
\bibfield{author}{\bibinfo{person}{Kannan Sridharan} {and} \bibinfo{person}{Gowri Sivaramakrishnan}.} \bibinfo{year}{2024}\natexlab{}.
\newblock \showarticletitle{Unlocking the potential of advanced large language models in medication review and reconciliation: A proof-of-concept investigation}.
\newblock \bibinfo{journal}{\emph{Exploratory Research in Clinical and Social Pharmacy}}  \bibinfo{volume}{15} (\bibinfo{year}{2024}), \bibinfo{pages}{100492}.
\newblock


\bibitem[Thirunavukarasu et~al\mbox{.}(2023)]%
        {thirunavukarasu2023large}
\bibfield{author}{\bibinfo{person}{Arun~James Thirunavukarasu}, \bibinfo{person}{Darren Shu~Jeng Ting}, \bibinfo{person}{Kabilan Elangovan}, \bibinfo{person}{Laura Gutierrez}, \bibinfo{person}{Ting~Fang Tan}, {and} \bibinfo{person}{Daniel Shu~Wei Ting}.} \bibinfo{year}{2023}\natexlab{}.
\newblock \showarticletitle{Large language models in medicine}.
\newblock \bibinfo{journal}{\emph{Nature Medicine}} \bibinfo{volume}{29}, \bibinfo{number}{8} (\bibinfo{year}{2023}), \bibinfo{pages}{1930--1940}.
\newblock


\bibitem[Uzuner et~al\mbox{.}(2010)]%
        {uzuner2010extracting}
\bibfield{author}{\bibinfo{person}{{\"O}zlem Uzuner}, \bibinfo{person}{Imre Solti}, {and} \bibinfo{person}{Eithon Cadag}.} \bibinfo{year}{2010}\natexlab{}.
\newblock \showarticletitle{Extracting medication information from clinical text}.
\newblock \bibinfo{journal}{\emph{Journal of the American Medical Informatics Association}} \bibinfo{volume}{17}, \bibinfo{number}{5} (\bibinfo{year}{2010}), \bibinfo{pages}{514--518}.
\newblock


\bibitem[Vrgoc et~al\mbox{.}(2024)]%
        {vrgoc2024millenniumdb}
\bibfield{author}{\bibinfo{person}{Domagoj Vrgoc}, \bibinfo{person}{Carlos Rojas}, \bibinfo{person}{Renzo Angles}, \bibinfo{person}{Marcelo Arenas}, \bibinfo{person}{Vicente Calisto}, \bibinfo{person}{Benjam{\'\i}n Far{\'\i}as}, \bibinfo{person}{Sebast{\'\i}an Ferrada}, \bibinfo{person}{Tristan Heuer}, \bibinfo{person}{Aidan Hogan}, \bibinfo{person}{Gonzalo Navarro}, {et~al\mbox{.}}} \bibinfo{year}{2024}\natexlab{}.
\newblock \showarticletitle{Millennium{DB}: A multi-modal, multi-model graph database}. In \bibinfo{booktitle}{\emph{Companion of the 2024 International Conference on Management of Data}}. \bibinfo{pages}{496--499}.
\newblock


\bibitem[{World Health Organization}(2019)]%
        {icd_2019}
\bibfield{author}{\bibinfo{person}{{World Health Organization}}.} \bibinfo{year}{2019}\natexlab{}.
\newblock \bibinfo{title}{International Classification of Diseases, 10th Edition}.
\newblock \bibinfo{howpublished}{\url{https://icd.who.int/browse10/2019/en}}.
\newblock
\newblock
\shownote{Accessed: 2026-1-16}.


\bibitem[Wu et~al\mbox{.}(2024)]%
        {wu2024pmc}
\bibfield{author}{\bibinfo{person}{Chaoyi Wu}, \bibinfo{person}{Weixiong Lin}, \bibinfo{person}{Xiaoman Zhang}, \bibinfo{person}{Ya Zhang}, \bibinfo{person}{Weidi Xie}, {and} \bibinfo{person}{Yanfeng Wang}.} \bibinfo{year}{2024}\natexlab{}.
\newblock \showarticletitle{{PMC-LLaMA}: Toward building open-source language models for medicine}.
\newblock \bibinfo{journal}{\emph{Journal of the American Medical Informatics Association}} \bibinfo{volume}{31}, \bibinfo{number}{9} (\bibinfo{year}{2024}), \bibinfo{pages}{1833--1843}.
\newblock


\bibitem[Wu et~al\mbox{.}(2025)]%
        {wu2025medical}
\bibfield{author}{\bibinfo{person}{Junde Wu}, \bibinfo{person}{Jiayuan Zhu}, \bibinfo{person}{Yunli Qi}, \bibinfo{person}{Jingkun Chen}, \bibinfo{person}{Min Xu}, \bibinfo{person}{Filippo Menolascina}, \bibinfo{person}{Yueming Jin}, {and} \bibinfo{person}{Vicente Grau}.} \bibinfo{year}{2025}\natexlab{}.
\newblock \showarticletitle{Medical graph {RAG}: Evidence-based medical large language model via graph retrieval-augmented generation}. In \bibinfo{booktitle}{\emph{Proceedings of the 63rd Annual Meeting of the Association for Computational Linguistics (Volume 1: Long Papers)}}. \bibinfo{pages}{28443--28467}.
\newblock


\bibitem[Xiao et~al\mbox{.}(2019)]%
        {xiao2019virtual}
\bibfield{author}{\bibinfo{person}{Guohui Xiao}, \bibinfo{person}{Linfang Ding}, \bibinfo{person}{Benjamin Cogrel}, {and} \bibinfo{person}{Diego Calvanese}.} \bibinfo{year}{2019}\natexlab{}.
\newblock \showarticletitle{Virtual knowledge graphs: An overview of systems and use cases}.
\newblock \bibinfo{journal}{\emph{Data Intelligence}} \bibinfo{volume}{1}, \bibinfo{number}{3} (\bibinfo{year}{2019}), \bibinfo{pages}{201--223}.
\newblock


\bibitem[Xiao et~al\mbox{.}(2025)]%
        {xiao2025llm4vkg}
\bibfield{author}{\bibinfo{person}{Guohui Xiao}, \bibinfo{person}{Lin Ren}, \bibinfo{person}{Guilin Qi}, \bibinfo{person}{Haohan Xue}, \bibinfo{person}{MD Panfilo}, {and} \bibinfo{person}{Davide Lanti}.} \bibinfo{year}{2025}\natexlab{}.
\newblock \showarticletitle{{LLM4VKG}: Leveraging large language models for virtual knowledge graph construction}. In \bibinfo{booktitle}{\emph{Proceedings of the 34th International Joint Conference on Artificial Intelligence (IJCAI)}}.
\newblock


\bibitem[Yang et~al\mbox{.}(2025)]%
        {yang2025qwen3}
\bibfield{author}{\bibinfo{person}{An Yang}, \bibinfo{person}{Anfeng Li}, \bibinfo{person}{Baosong Yang}, \bibinfo{person}{Beichen Zhang}, \bibinfo{person}{Binyuan Hui}, \bibinfo{person}{Bo Zheng}, \bibinfo{person}{Bowen Yu}, \bibinfo{person}{Chang Gao}, \bibinfo{person}{Chengen Huang}, \bibinfo{person}{Chenxu Lv}, {et~al\mbox{.}}} \bibinfo{year}{2025}\natexlab{}.
\newblock \showarticletitle{Qwen3 technical report}.
\newblock \bibinfo{journal}{\emph{arXiv preprint arXiv:2505.09388}} (\bibinfo{year}{2025}).
\newblock


\bibitem[Zhang and Soh(2024)]%
        {zhang2024extract}
\bibfield{author}{\bibinfo{person}{Bowen Zhang} {and} \bibinfo{person}{Harold Soh}.} \bibinfo{year}{2024}\natexlab{}.
\newblock \showarticletitle{Extract, define, canonicalize: An {LLM}-based framework for knowledge graph construction}. In \bibinfo{booktitle}{\emph{Proceedings of the 2024 Conference on Empirical Methods in Natural Language Processing}}. \bibinfo{pages}{9820--9836}.
\newblock


\bibitem[Zhang et~al\mbox{.}(2025)]%
        {zhang2025siren}
\bibfield{author}{\bibinfo{person}{Yue Zhang}, \bibinfo{person}{Yafu Li}, \bibinfo{person}{Leyang Cui}, \bibinfo{person}{Deng Cai}, \bibinfo{person}{Lemao Liu}, \bibinfo{person}{Tingchen Fu}, \bibinfo{person}{Xinting Huang}, \bibinfo{person}{Enbo Zhao}, \bibinfo{person}{Yu Zhang}, \bibinfo{person}{Yulong Chen}, {et~al\mbox{.}}} \bibinfo{year}{2025}\natexlab{}.
\newblock \showarticletitle{Siren’s song in the {AI} ocean: A survey on hallucination in large language models}.
\newblock \bibinfo{journal}{\emph{Computational Linguistics}} (\bibinfo{year}{2025}), \bibinfo{pages}{1--46}.
\newblock


\bibitem[Zhao et~al\mbox{.}(2025)]%
        {zhao2024hybrid}
\bibfield{author}{\bibinfo{person}{Fuheng Zhao}, \bibinfo{person}{Divyakant Agrawal}, {and} \bibinfo{person}{Amr~El Abbadi}.} \bibinfo{year}{2025}\natexlab{}.
\newblock \showarticletitle{Hybrid querying over relational databases and large language models}. In \bibinfo{booktitle}{\emph{15th Annual Conference on Innovative Data Systems Research}}.
\newblock


\bibitem[Zhu et~al\mbox{.}(2024)]%
        {zhu2024llms}
\bibfield{author}{\bibinfo{person}{Yuqi Zhu}, \bibinfo{person}{Xiaohan Wang}, \bibinfo{person}{Jing Chen}, \bibinfo{person}{Shuofei Qiao}, \bibinfo{person}{Yixin Ou}, \bibinfo{person}{Yunzhi Yao}, \bibinfo{person}{Shumin Deng}, \bibinfo{person}{Huajun Chen}, {and} \bibinfo{person}{Ningyu Zhang}.} \bibinfo{year}{2024}\natexlab{}.
\newblock \showarticletitle{{LLM}s for knowledge graph construction and reasoning: Recent capabilities and future opportunities}.
\newblock \bibinfo{journal}{\emph{World Wide Web}} \bibinfo{volume}{27}, \bibinfo{number}{5} (\bibinfo{year}{2024}), \bibinfo{pages}{58}.
\newblock


\bibitem[Zitnik et~al\mbox{.}(2018)]%
        {zitnik2018modeling}
\bibfield{author}{\bibinfo{person}{Marinka Zitnik}, \bibinfo{person}{Monica Agrawal}, {and} \bibinfo{person}{Jure Leskovec}.} \bibinfo{year}{2018}\natexlab{}.
\newblock \showarticletitle{Modeling polypharmacy side effects with graph convolutional networks}.
\newblock \bibinfo{journal}{\emph{Bioinformatics}} \bibinfo{volume}{34}, \bibinfo{number}{13} (\bibinfo{year}{2018}), \bibinfo{pages}{i457--i466}.
\newblock


\end{thebibliography}

\end{document}